\documentclass{article}

\usepackage{tabularx}
\usepackage{graphicx}
\usepackage{multirow}
\usepackage{amssymb}
\usepackage{xcolor}
\usepackage{amsmath}
\usepackage{wrapfig}
\usepackage{algorithmic}
\usepackage{algorithm}
\usepackage{hyperref}

\newcommand{\improve}[2]{#1 {\scriptsize\textcolor{blue}{+#2}}}
\newcommand{\noimp}[2]{#1 {\scriptsize{+#2}}}
\newcommand{\decline}[2]{#1 {\scriptsize\textcolor{red}{-#2}}}

\usepackage[preprint]{neurips_2025}

\usepackage[utf8]{inputenc} 
\usepackage[T1]{fontenc}    
\usepackage{hyperref}       
\usepackage{url}            
\usepackage{booktabs}       
\usepackage{amsfonts}       
\usepackage{nicefrac}       
\usepackage{microtype}      
\usepackage{xcolor}         
\usepackage{soul}

\title{Code-Driven Planning in Grid Worlds with Large Language Models}

\author{%
  Ashwath Vaithinathan Aravindan \\
  University of Southern California \\
  United States of America \\
  \texttt{vaithina@usc.edu} \\
  \And
  Zhisheng Tang \\
  University of Southern California \\
  Information Sciences Institute \\
  Marina del Rey, 90292 \\
  United States of America \\
  \texttt{zhisheng@isi.edu} \\
  \AND
  Mayank Kejriwal \\
  University of Southern California \\
  Information Sciences Institute \\
  Marina del Rey, 90292 \\
  United States of America \\
  \texttt{kejriwal@isi.edu} \\
}

\begin{document}

\maketitle

\begin{abstract}

We propose an iterative programmatic planning (IPP) framework for solving grid-based tasks by synthesizing interpretable agent policies expressed in code using large language models (LLMs). Instead of relying on traditional search or reinforcement learning, our approach uses code generation as policy synthesis, where the LLM outputs executable programs that map environment states to action sequences. Our proposed architecture incorporates several prompting strategies, including direct code generation, pseudocode-conditioned refinement, and curriculum-based prompting, but also includes an iterative refinement mechanism that updates code based on task performance feedback. We evaluate our approach using six leading LLMs and two challenging grid-based benchmarks (GRASP and MiniGrid). Our IPP framework demonstrates improvements over direct code generation ranging from 10\% to as much as 10x across five of the six models and establishes a new state-of-the-art result for GRASP. IPP is found to significantly outperform direct elicitation of a solution from GPT-o3-mini (by 63\% on MiniGrid to 116\% on GRASP), demonstrating the viability of the overall approach. Computational costs of all code generation approaches are similar. While code generation has a higher initial prompting cost compared to direct solution elicitation (\$0.08 per task vs. \$0.002 per instance for GPT-o3-mini), the code can be reused for any number of instances, making the amortized cost significantly lower (by 400x on GPT-o3-mini across the complete GRASP benchmark).

\end{abstract}

\section{Introduction}

Grid-based environments are a core abstraction for planning tasks in robotics \cite{zholus2022iglu, ajeil2020grid, tripathy2021care}, embodied AI \cite{jain2021gridtopix}, and reinforcement learning (RL) \cite{chevalier2023minigrid, chevalier2018babyai, MinigridMiniworld23, 6907756, eberhardinger2023learning}. These environments offer a formal, discrete setting to study core aspects of intelligent behavior, such as planning \cite{tamar2016value, sukhbaatar2015mazebase}, exploration \cite{pathak2017curiosity}, and generalization \cite{cobbe2020leveraging}. Their simplicity enables systematic evaluation across diverse approaches: from classical search \cite{Hart1968} and reinforcement learning \cite{watkins1992q, metropolis1949monte} to neuro-symbolic \cite{hitzler2022neuro} and language-based methods \cite{topsakal2024evaluating}. They also abstract real-world domains like robotics and navigation, capturing spatial and symbolic reasoning while removing perception noise and sensor complexity. This makes them an ideal testbed for studying decision-making and planning, linking high-level reasoning to low-level control.

Meanwhile, large language models (LLMs) have recently shown strong performance on a variety of reasoning tasks, including question answering \cite{kamalloo2023evaluating, singhal2025toward}, tool use \cite{qu2025tool, qin2023toolllm}, and code generation \cite{jiang2024survey, roziere2023code}. However, LLMs have struggled with planning \cite{valmeekam2023planning}, which requires state tracking, goal-directed search, and consistent multi-step decision-making, all of which LLMs do not naturally exhibit. At the same time, LLMs excel at generating interpretable code \cite{jiang2024survey}, which motivates a shift in perspective: instead of prompting LLMs to produce action plans directly, we treat planning as code generation. LLMs can generate executable programs that act as action policies. This capability aligns \textit{planning with programming}, since both require symbolic reasoning over structured inputs and constraints. Additionally, such an approach offers several advantages: generated programs are interpretable, reusable across task instances, and cheap to execute. 

\begin{figure}
    \centering
    \includegraphics[width=\linewidth]{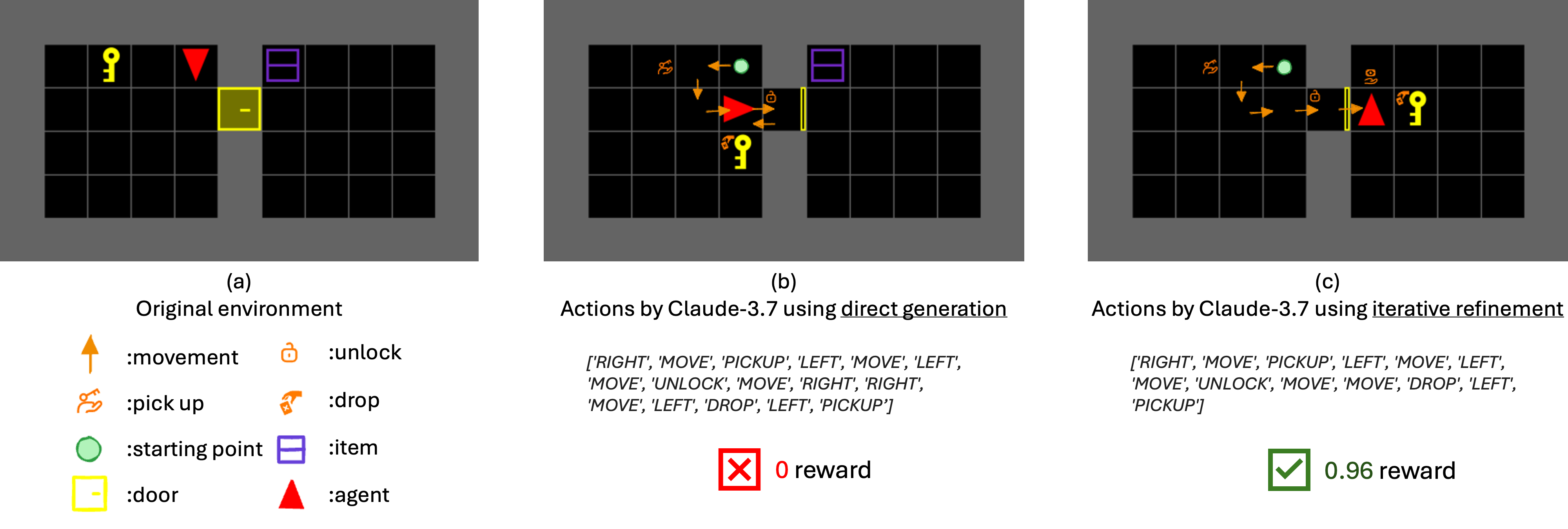}
    \caption{Example of the \textit{Unlock Pickup} task from the MiniGrid benchmark. The agent (red triangle) must retrieve a box (purple item) located in a separate room behind a locked door (yellow). Directions in the action sequence indicate the direction the agent turns to face before moving. (a) shows the original environment setup (b) shows actions generated by Claude-3.7 using \textit{direct generation}, which fails and yields a reward of 0. (c) shows \textit{iterative refinement}, which successfully completes the task with a reward of 0.96.}
    \label{fig:example}
\end{figure}

Compared to RL, which relies on trial-and-error interaction and costly gradient-based optimization, a code-driven approach offers a lightweight and interpretable alternative that requires no task-specific training. While RL excels in grid-based domains, it often suffers from poor generalization \cite{cobbe2019quantifying, korkmaz2024survey} and produces opaque policies \cite{eberhardinger2023learning, cheng2025survey}. In contrast, a code-driven approach replaces policy learning with LLM-based program synthesis, enabling the generation of symbolic and reusable policies that are efficient and transparent, and aligning with a broader shift toward grounding LLMs in executable environments \cite{liang2023code, xu2023creative, wang2024executable} where code serves as a verifiable and editable interface between high-level reasoning and low-level control. If generated code can be demonstrated to perform sufficiently well on such tasks, it substitutes for RL as a practical and scalable solution for building generalizable planning agents. 

In this work, we propose \textbf{Iterative Programmatic Planning (IPP)}, a code-driven framework that uses LLMs to generate Python programs as action policies. IPP incorporates an \textbf{Iterative Refinement (IR)} loop, where LLM-generated programs are executed, evaluated, and revised based on feedback to correct errors and improve task performance. Figure~\ref{fig:example} illustrates one example of performance improvement enabled by iterative refinement. We evaluate IPP on two grid-based benchmarks: GRASP \cite{tang2024grasp} and MiniGrid \cite{MinigridMiniworld23}, across six state-of-the-art LLMs, and show that it produces robust and generalizable \textit{code-based policies}. Compared to directly prompting for action sequence and other baselines, IPP offers better performance, greater transparency, and lower computational cost. In summary, this work makes the following contributions:

\begin{itemize}
    \item We propose Iterative Programmatic Planning (IPP), a novel framework that treats planning as code generation. Instead of producing action sequences directly, LLMs synthesize interpretable Python programs as reusable policies for grid-based environments. IPP introduces an \textit{Iterative Refinement} process that improves these programs based on task-specific feedback. This approach offers transparency, efficiency, and generalization without requiring task-specific training.
    
    \item We evaluate IPP on two benchmarks: GRASP and MiniGrid, across six leading LLMs. IPP improves over direct code generation by 10\% to 10× on GRASP and raises MiniGrid completion rates from 0\% to 84\% for GPT-o3-mini. Compared to non-code-based CoT and 2-step CoT prompting, IPP attains equal or better performance at significantly lower cost, since its synthesized policies generalize without requiring per-instance inference.

    \item We introduce two structured prompting strategies: \textit{pseudocode Extension}, where LLMs turn a high-level algorithmic sketch into executable code, and \textit{step-by-step curriculum prompting}, where the complexity of the task is incrementally added. In the GRASP benchmark, pseudocode extension improves energy collection for Claude-3.7 from 0.89 (Greedy) to 0.98 (+10\%), and for GPT-o3-mini to 1.36 (+52\%). Using step-by-step curriculum prompting, GPT-o3-mini achieves steady gains --- from 0.44 to 1.73 (+293\%) at final complexity --- while Claude-3.7 improves from 0.12 to 1.42 (+1083\%) from Step 0 to Step 2 before leveling off, demonstrating the benefit of curriculum-based synthesis.
    
\end{itemize}

\section{Approach}

We consider planning problems that can be represented as discrete, grid-based environments. We assume that each environment is fully observable and composed of cells, and each cell is associated with attributes such as obstacles, items, and goal-relevant entities. Agents must navigate these grids and execute a task e.g., item collection or goal-reaching—by performing valid action sequences that satisfy task-specific constraints. These environments are common in domains like embodied AI, robotics, and reinforcement learning \cite{MinigridMiniworld23, jain2021gridtopix, tripathy2021care}.

Formally, each planning problem is represented as a tuple $\mathcal{P} = (\mathcal{S}, \mathcal{A}, T, s_0, \mathcal{G})$, where $\mathcal{S}$ is the state space; each $s \in \mathcal{S}$ denotes a configuration of the grid environment, including the agent's position and attributes (e.g., energy, inventory), as well as locations of objects and obstacles. $\mathcal{A}$ is the finite set of discrete actions (e.g., move, pick up, unlock, drop) that the agent is allowed. $T: \mathcal{S} \times \mathcal{A} \rightarrow \mathcal{S}$ is the pre-defined transition function. We assume the transition function is deterministic and fully defined, so the effects of each action in a given state are predictable and known. $s_0 \in \mathcal{S}$ is the initial state. We assume that this is fully observable, meaning the agent has access to the complete initial state of the environment. Finally, $\mathcal{G} \subseteq \mathcal{S}$ is the set of goal states.

A valid plan is then defined as a sequence of actions $\pi = [a_0, a_1, \dots, a_T]$ such that the resulting state after executing $\pi$ from $s_0$ falls within the goal region: $T(T(\dots T(s_0, a_0), a_1), \dots, a_T) \in \mathcal{G}$. The agent's task is to generate a sequence of actions $\pi = [a_0, a_1, \dots, a_T]$ that transforms the initial state $s_0 \in \mathcal{S}$ into a goal state $s_T \in \mathcal{G}$ under the transition function $T$. This requires reasoning about the structure of the environment, anticipating the effects of actions via $T(s, a)$, and selecting actions from $\mathcal{A}$ that satisfy task-specific constraints while getting closer toward the objective.

To solve such planning problems, we propose an Iterative Programmatic Planning (IPP) framework that uses LLMs to synthesize interpretable and executable programs that serve as action policies. In this framework, the LLM generates programs that map environment observations to sequences of allowable actions, effectively functioning as planning agents. We first decompose the problem into three key components: (\textit{1) direct program synthesis}, where the LLM is prompted to generate complete, executable programs in a general-purpose interpretable language (e.g., Python) from natural language task descriptions; \textit{(2) pseudocode-conditioned refinement}, where the model is seeded with a high-level algorithmic sketch (e.g., greedy search) and tasked with completing or adapting it to the task environment; and \textit{(3) curriculum-based prompting}, where the model is guided through a sequence of incrementally complex subproblems to facilitate generalization to harder tasks. 

We combine these components into a unified prompting-based approach suitable for zero-shot and few-shot planning through code generation. However, to further \textit{optimize} the program generated by the model, we also incorporate an iterative refinement process whereby the LLM first generates an initial candidate program, which is then executed and evaluated on a set of task instances. Based on some performance metric (such as reward and completion rate), the model is prompted to revise the code. Iteration continues until performance converges or improves no further. Through refinement, we show that the LLM learns to correct logical errors, incorporate missing constraints, and restructure control flow.

\subsection{Code-driven Planning with Iterative Refinement}

Rather than relying on traditional search algorithms or reinforcement learning to solve $\mathcal{P}$, code-as-plan frameworks use an LLM to generate an interpretable program $f$ that, when executed against a specific environment instance, produces a sequence of actions $\pi = [a_0, a_1, \dots, a_T]$ transitioning the agent from the initial state $s_0$ to a goal state $s_T \in \mathcal{G}$. This program acts as a reusable and interpretable policy that maps environment observations to action sequences.

This approach contrasts with classical reinforcement learning methods, which typically learn a policy $\pi: \mathcal{S} \rightarrow \mathcal{A}$ through repeated interaction with the environment and optimization of cumulative rewards, a process that is time-consuming and computationally intensive. At the other extreme, an LLM could be prompted to directly produce a plan given a grid-based planning problem. Experimentally, this approach is found to yield significantly low-quality solutions, even for leading LLMs (Section \ref{sec:results}; see also Figure \ref{fig:example}). Code-driven planning offers a compromise by forcing the LLM to provide an explicit policy, using only a few prompts, encoded as an interpretable program. An auxiliary advantage of such a policy is that its logic is observable and interpretable.

\paragraph{Direct Generation}  
Our baseline approach, \emph{direct generation}, prompts the LLM to directly synthesize a complete and executable program $f$ given only a natural language task description and environment specifications. This approach assumes that the model can generate code that considers both a high-level strategy and low-level execution details in a single pass. While fast and straightforward, this approach may overlook edge cases or misapply rules. Nonetheless, it establishes a strong starting point for comparing more advanced techniques\footnote{All prompt templates for the steps described in this section are provided in Appendices \ref{app:grasp_prompt}, \ref{app:door_key_prompt}, \ref{app:unlock_pickup_prompt}, and \ref{app:unlock_prompt}.}.

\paragraph{Pseudocode Extension}  
To address some of the flaws of direct generation, we introduce a pseudocode-based prompting strategy. Here, the LLM is provided with a high-level baseline method (e.g., greedy or rule-based logic in pseudocode) and tasked with extending and refining it into an executable program $f$. This approach narrows the LLM’s search space, allowing the model to use \textit{symbolic priors} while preserving flexibility. A disadvantage of this approach is that it requires the pseudocode prior to be provided in the first place. For complex grids, providing such a prior may be cumbersome or even infeasible. 

\paragraph{Step-by-Step}  
The \emph{step-by-step} approach introduces a curriculum-style procedure, in which the model is prompted to solve increasingly complex versions of the task, gradually building up to the full problem. Beginning with simplified environments (e.g., no obstacles, no step cost), the model produces initial code $f^{(0)}$ that works under minimal assumptions. At each subsequent stage $i$, the model is given an updated problem with more complexity, such as limited energy, obstacles, or carry constraints, and is required to revise and extend the previous program $f^{(i-1)}$ into a new program $f^{(i)}$ that can solve the current problem. This formulation mimics human incremental reasoning and in theory, should help the LLM produce a strategy that it might fail to produce through (for example) one-pass direct generation. 

\begin{wrapfigure}{r}{0.55\textwidth}
\vspace{-1em}
\begin{minipage}{\linewidth}
\scriptsize 
\textbf{Algorithm 2.2: Iterative Refinement}

\begin{algorithmic}[1]\label{alg:refinement}
\REQUIRE Planning problem $\mathcal{P} = (\mathcal{S}, \mathcal{A}, T, s_0, \mathcal{G})$
\STATE $f^{(0)} \leftarrow$ \texttt{DirectGeneration}$(\mathcal{P})$, \quad $J_{\text{prev}} \leftarrow -\infty$
\STATE $t \leftarrow 0$
\WHILE{True}
    \STATE $\mathcal{D} \leftarrow$ evaluation results of $f^{(t)}$ on training set
    \STATE $J_{\text{curr}} \leftarrow$ average over all $J^{(i)}$ in $\mathcal{D}$
    \IF{$J_{\text{curr}} \leq J_{\text{prev}}$}
        \STATE \textbf{break}
    \ENDIF
    \STATE $J_{\text{prev}} \leftarrow J_{\text{curr}}$
    \STATE $\mathcal{D}_{\text{fail}} \leftarrow \texttt{SelectWorstCases}(\mathcal{D}, k=3)$
    \STATE $f^{(t+1)} \leftarrow \texttt{RefineWithFeedback}(f^{(t)}, \mathcal{D}_{\text{fail}})$
    \STATE $t \leftarrow t + 1$
\ENDWHILE
\STATE \RETURN $f^{(t)}$
\end{algorithmic}
\end{minipage}
\vspace{-2em}
\end{wrapfigure}

\paragraph{Iterative Refinement}
To improve the correctness and generality of the generated program $f$, we propose an \textit{Iterative Refinement (IR)} strategy, whereby an LLM iteratively improves its generated code through alternating steps of evaluation and feedback-driven refinement (Algorithm~\ref{alg:refinement}). Starting from an initial candidate program $f^{(0)}$ (obtained via direct generation), we evaluate its performance on a training set of planning problems $\mathcal{P}$ and measure its performance via a pre-defined task-specific metric $J^{(i)}$ for each instance.

The aggregate performance $J_{\text{curr}} = \texttt{Aggregate}({J^{(i)}})$ is then compared to the previous iteration’s performance $J_{\text{prev}}$. If no improvement is observed, we assume convergence and terminate the loop. Otherwise, we identify the $k$ worst-performing instances (i.e., those with the lowest $J^{(i)}$ values) as $\mathcal{D}_{\text{fail}}$, and prompt the LLM to refine the current program $f^{(t)}$ based on this feedback.  The revised program $f^{(t+1)}$ is expected to address the identified issues while preserving correct behavior elsewhere.

From an optimization perspective, IR can be viewed as a discrete, non-gradient-based search procedure over the space of executable programs. Each refinement step attempts to improve a program $f^{(t)}$ by minimizing an implicit loss function, which is captured via task-specific evaluation metrics, over a finite dataset of grid instances. The feedback-driven update acts as a heuristic gradient, guiding the LLM to generate a new program $f^{(t+1)}$ that corrects previous failures while maintaining performance on successful instances. While analytically not an optimizer, IR approximates iterative function minimization in a symbolic space, guided by performance signals rather than gradients.

\section{Experimental setup}
We evaluate our proposed framework on two benchmarks: the GRASP dataset \cite{tang2024grasp} and three tasks from MiniGrid \cite{MinigridMiniworld23}. They are chosen for their structured 2D grid environments, which can be easily converted into LLM prompts (Appendix~\ref{app:prompts}).

\paragraph{Evaluation Benchmarks} The GRASP benchmark is a spatial planning task that requires an agent to collect energy under different constraints. Each environment instance varies by movement type, carry limit, per-step energy cost, energy distribution, obstacles, and starting position, yielding 16,000 unique instances. Performance is measured by average energy collected within 20 steps, implicitly penalizing inefficient movement and suboptimal planning. In parallel, we evaluate the LLMs on three tasks from the MiniGrid benchmark: \textit{Unlock, Door-Key}, and \textit{Unlock-Pickup}. MiniGrid is a lightweight grid-based benchmark for spatial reasoning, navigation, and object interaction. The selected three tasks increase in complexity. For each task, we generate 1,000 instances and measure two metrics: the average reward and the average completion rate.

\paragraph{Prompting Strategies and Baselines} We evaluate the IPP framework using four prompting strategies: \textit{Direct Generation, Pseudocode Extension, Step-by-Step,} and \textit{Iterative Refinement}. We also compare it to four non-code-based planning and prompting baselines. Using the Chain-of-Thought (CoT) methods \cite{wei2022chain}, the model is prompted to reason through intermediate steps in natural language before outputting the full action plan. We also introduce a 2-Step CoT variant that splits the planning into two stages: the model first generates a partial action plan up to a defined midpoint, then completes it based on the updated environment. This approach reduces planning burden by dividing the task, enabling more focused and accurate reasoning at each stage. 
Finally, we evaluate two simple programmatic baselines: \textit{Random} and \textit{Greedy}. The \textit{Random} agent selects valid actions uniformly at random, serving as a lower bound. The \textit{Greedy} agent follows a hand-coded, myopic policy based on immediate utility, offering a fast but potentially short-sighted comparison.

\paragraph{LLMs Evaluated} We evaluate our IPP framework using six LLMs, including both general-purpose LLMs and code-specialized ones. Specifically, we use the following: GPT-4o \cite{openai2024gpt4o}, GPT-o1 \cite{openai2024gpto1}, GPT-o3-mini \cite{openai2025gpto3mini}, Claude 3.7 Sonnet \cite{anthropic2025claude37}, Gemini 2.5 Pro \cite{google2025gemini25pro}, and DeepSeek-R1 \cite{deepseek2025r1}. All models are accessed through the OpenRouter API \cite{openrouter2024}. Code for reproducing all experiments is available \href{https://drive.google.com/file/d/1SQEtcloi0U_0DyO9ecU5NtjwnFO-wPP0/view?usp=sharing}{here}.

\section{Results}\label{sec:results}

Table~\ref{tab:grasp_results_obstacle} shows the average energy collected across five energy distributions, with and without obstacles, in the GRASP dataset, comparing Direct Generation (DG) with Iterative Refinement (IR). Across most models (except for DeepSeek-R1) and environment configurations, IR consistently outperforms DG. Overall, GPT-o1 and Claude-3.7 show substantial gains, with IR achieving up to 429\% and 143\% improvement, particularly in cluster and skewed energy distributions. Gemini-2.5-Pro exhibits the largest percentage improvement overall, turning negative DG scores (e.g., –0.48) into high-positive IR scores (e.g., 3.27), with gains up to 1011\%. GPT-o3-mini, despite strong baseline performance in DG (e.g., 2.9 average energy in obstacle-free environments), benefits only slightly from IR (1 to 8\% gains). In contrast, GPT-4o shows limited effectiveness, with net energy scores remaining negative even after IR, while DeepSeek-R1 uniquely exhibits consistent performance declines (up to –20\%) across all settings, suggesting potential incompatibility with the IR strategy.

\begin{table}
\centering
\renewcommand{\arraystretch}{1.6}
\caption{Performance of LLMs on the GRASP benchmark under different energy distributions (Random, Vertically-skewed, Horizontally-skewed, Cluster, Spiral) with and without obstacles. Two prompting strategies are compared: \textbf{Direct Generation (DG)} and \textbf{Iterative Refinement (IR)}. The numbers represent the average net energy collected; blue/red text indicates the percentage improvement or deterioration of IR over DG.}
\label{tab:grasp_results_obstacle}
\resizebox{\textwidth}{!}{%
\begin{tabular}{llcccccccccccc}
\toprule
\multirow{2}{*}{\textbf{Model}}  & \multirow{2}{*}{\textbf{Obstacle}} 
& \multicolumn{2}{c}{\textbf{Random}} 
& \multicolumn{2}{c}{\textbf{V-skewed}} 
& \multicolumn{2}{c}{\textbf{H-skewed}} 
& \multicolumn{2}{c}{\textbf{Cluster}} 
& \multicolumn{2}{c}{\textbf{Spiral}} \\
\cmidrule(r){3-4} \cmidrule(r){5-6} \cmidrule(r){7-8} \cmidrule(r){9-10} \cmidrule(r){11-12}
 & 
& DG & IR 
& DG & IR 
& DG & IR 
& DG & IR 
& DG & IR \\
\midrule
\multirow{2}{*}{GPT-4o}         
& With    & -1.40 & \improve{-0.38}{73\%} & -1.46 & \improve{-0.40}{73\%} & -1.47 & \improve{-0.40}{73\%} & -1.93 & \improve{-0.62 }{68\%} & -1.62 & \improve{-0.46 }{72\%} \\

& Without & -1.32 & \improve{-0.36}{73\%} & -1.35 & \improve{-0.34}{75\%} & -1.26 & \improve{-0.34}{73\%} & -1.96 & \improve{-0.69}{65\%} & -1.56 & \improve{-0.47}{70\%} \\

\addlinespace
\multirow{2}{*}{GPT-o1}         
& With    & 1.09 & \improve{2.89}{165\%} & 1.04 & \improve{2.85}{174\%} & 1.01 & \improve{2.86}{183\%} & 0.50 & \improve{2.51}{402\%} & 0.81 & \improve{2.60}{221\%} \\

& Without & 1.16 & \improve{3.04}{162\%} & 1.18 & \improve{3.03}{157\%} & 1.19 & \improve{3.10}{161\%} & 0.49 & \improve{2.59}{429\%} & 0.90 & \improve{2.76}{207\%} \\

\addlinespace
\multirow{2}{*}{GPT-o3-mini}    
& With    & 2.89 & \improve{2.92}{1\%} & 2.85 & \improve{2.92}{2\%} & 2.86 & \improve{2.91}{2\%} & 2.51 & \improve{2.72}{8\%} & 2.60 & \improve{2.71}{4\%} \\

& Without & 3.04 & \noimp{3.04}{0\%} & 3.03 & \improve{3.05}{1\%} & 3.10 & \decline{3.07}{1\%} & 2.59 & \improve{2.78}{7\%} & 2.76 & \improve{2.85}{3\%} \\

\addlinespace
\multirow{2}{*}{Claude-3.7}     
& With    & 1.32 & \improve{2.82}{114\%} & 1.19 & \improve{2.79}{134\%} & 1.15 & \improve{2.79}{143\%} & 1.30 & \improve{2.60}{100\%} & 1.33 & \improve{2.60}{95\%} \\

& Without & 1.32 & \improve{2.92}{121\%} & 1.20 & \improve{2.91}{143\%} & 1.33 & \improve{2.96}{123\%} & 1.51 & \improve{2.66}{76\%} & 1.31 & \improve{2.76}{111\%} \\

\addlinespace
\multirow{2}{*}{Gemini-2.5-Pro} 
& With    & -0.48 & \improve{3.27}{781\%} & -0.45 & \improve{3.28}{829\%} & -0.50 & \improve{3.26}{752\%} & -0.80 & \improve{3.00}{475\%} & -0.54 & \improve{2.96}{648\%} \\

& Without & -0.44 & \improve{3.38}{868\%} & -0.41 & \improve{3.39}{927\%} & -0.38 & \improve{3.46}{1011\%} & -0.79 & \improve{3.09}{491\%} & -0.46 & \improve{3.12}{778\%} \\

\addlinespace
\multirow{2}{*}{DeepSeek-R1}    
& With    & 1.96 & \decline{1.65}{16\%} & 1.87 & \decline{1.63}{13\%} & 1.85 & \decline{1.62}{12\%} & 1.66 & \decline{1.40}{16\%} & 1.62 & \decline{1.55}{4\%} \\

& Without & 2.00 & \decline{1.68}{16\%} & 2.02 & \decline{1.68}{17\%} & 2.06 & \decline{1.69}{18\%} & 1.73 & \decline{1.38}{20\%} & 1.78 & \decline{1.59}{11\%} \\
\bottomrule
\end{tabular}
}
\end{table}

Table~\ref{tab:minigrid_reward} presents the performance of LLMs on three MiniGrid tasks: Unlock, Door Key, and Unlock Pickup, comparing the DG and IR strategies using two evaluation metrics: average reward and completion rate. Among all models, GPT-o1 consistently shows perfect or near-perfect performance under DG, leaving no room for improvement for IR. In contrast, GPT-o3-mini and Claude-3.7 benefit substantially from IR, especially in more complex tasks like Unlock-Pickup. GPT-o3-mini improves from total failure (0.00 reward, 0\% completion) to strong performance (0.78 reward, 84\% completion), while Claude-3.7 achieves similar gains, reaching 1.00 completion and a 70\% increase in reward.

Other models show more varied results. Gemini-2.5-Pro displays meaningful improvement only in the Door-Key task, with static performance elsewhere. DeepSeek-R1 shows inconsistent behavior: IR dramatically increases performance in the Unlock task from 0.00 to 0.97 in reward, yet decreases performance in Door-Key and has no gain on Unlock-Pickup. On the other hand, GPT-4o fails to improve using IR across all tasks, maintaining 0 reward and completion rate, indicating fundamental issues with task comprehension or execution.

\begin{table}
\centering
\renewcommand{\arraystretch}{1.6}
\caption{Performance of LLMs on three MiniGrid tasks: \textbf{Unlock}, \textbf{Door-Key}, and \textbf{Unlock-Pickup}. The performance is evaluated using two metrics: \textbf{average reward} and \textbf{completion rate}, each reported as \textbf{mean ± standard deviation}. Each model is tested using two strategies: \textbf{Direct Generation (DG)} and \textbf{Iterative Refinement (IR)}. Blue/red text indicates the percentage improvement or deterioration of IR over DG.}
\label{tab:minigrid_reward}
\resizebox{\textwidth}{!}{%
\begin{tabular}{llccccccc}
\toprule
\multirow{2}{*}{\textbf{Model}} & \multirow{2}{*}{\textbf{Metric}} & \multicolumn{2}{c}{Unlock} & \multicolumn{2}{c}{Door-Key} & \multicolumn{2}{c}{Unlock-Pickup} \\
\cmidrule(lr){3-4} \cmidrule(lr){5-6} \cmidrule(lr){7-8}
& & DG & IR & DG & IR & DG & IR \\
\midrule
\multirow{2}{*}{GPT-4o} 
& Reward           & 0.00 $\pm$ 0.00 & \noimp{0.00 $\pm$ 0.00}{0\%} & 0.00 $\pm$ 0.00 & \noimp{0.00 $\pm$ 0.00}{0\%}     & 0.00 $\pm$ 0.00 & \noimp{0.00 $\pm$ 0.00}{0\%} \\
& Completion Rate  & 0.00 $\pm$ 0.00 & \noimp{0.00 $\pm$ 0.00}{0\%} & 0.00 $\pm$ 0.00 & \noimp{0.00 $\pm$ 0.00}{0\%} & 0.00 $\pm$ 0.00 & \noimp{0.00 $\pm$ 0.00}{0\%} \\
\multirow{2}{*}{GPT-o1} 
& Reward           & 0.97 $\pm$ 0.01 & \noimp{0.97 $\pm$ 0.01}{0\%}    & 0.97 $\pm$ 0.00 & \noimp{0.97 $\pm$ 0.00}{0\%}    & 0.95 $\pm$ 0.01  & \noimp{0.95 $\pm$ 0.01}{0\%}    \\
& Completion Rate  & 1.00 $\pm$ 0.00 & \noimp{1.00 $\pm$ 0.00}{0\%} & 1.00 $\pm$ 0.00 & \noimp{1.00 $\pm$ 0.00}{0\%} & 1.00 $\pm$ 0.00 & \noimp{1.00 $\pm$ 0.00}{0\%} \\
\multirow{2}{*}{GPT-o3-mini} 
& Reward           & 0.90 $\pm$ 0.24 & \improve{0.97 $\pm$ 0.01}{8\%}    & 0.78 $\pm$ 0.39 & \improve{0.98 $\pm$ 0.00}{26\%} & 0.00 $\pm$ 0.00 & \improve{0.78 $\pm$ 0.34}{$\infty$\%}    \\
& Completion Rate  & 0.93 $\pm$ 0.25 & \improve{1.00 $\pm$ 0.00}{8\%}    & 0.80 $\pm$ 0.40 & \improve{1.00 $\pm$ 0.00}{25\%}    & 0.00 $\pm$ 0.00 & \improve{0.84 $\pm$ 0.36}{$\infty$\%}    \\
\multirow{2}{*}{Claude-3.7} 
& Reward           & 0.71 $\pm$ 0.42 & \improve{0.90 $\pm$ 0.24}{27\%} & 0.75 $\pm$ 0.41 & \improve{0.88 $\pm$ 0.28}{17\%}& 0.55 $\pm$ 0.46 & \improve{0.94 $\pm$ 0.01}{70\%}    \\
& Completion Rate  & 0.74 $\pm$ 0.44 & \improve{0.93 $\pm$ 0.25}{27\%} & 0.78 $\pm$ 0.42 & \improve{0.91 $\pm$ 0.29}{17\%} & 0.59 $\pm$ 0.49 & \improve{1.00 $\pm$ 0.00}{69\%}    \\
\multirow{2}{*}{Gemini-2.5-Pro} 
& Reward           & 0.97 $\pm$ 0.01 & \noimp{0.97 $\pm$ 0.01}{0\%}    & 0.71 $\pm$ 0.43 & \improve{0.90 $\pm$ 0.26}{27\%}     & 0.89 $\pm$ 0.22 & \noimp{0.89 $\pm$ 0.22}{0\%}    \\
& Completion Rate  & 1.00 $\pm$ 0.00 & \noimp{1.00 $\pm$ 0.00}{0\%} & 0.73 $\pm$ 0.44 & \improve{0.92 $\pm$ 0.27}{26\%} & 0.91 $\pm$ 0.29 & \noimp{0.91 $\pm$ 0.29}{0\%} \\
\multirow{2}{*}{DeepSeek-R1} 
& Reward           & 0.00 $\pm$ 0.00 & \improve{0.97 $\pm$ 0.01}{$\infty$\%} & 0.92 $\pm$ 0.23 & \decline{0.83 $\pm$ 0.35}{10\%} & 0.86 $\pm$ 0.28 & \noimp{0.86 $\pm$ 0.28}{0\%}\\
& Completion Rate  & 0.00 $\pm$ 0.00 & \improve{1.00 $\pm$ 0.00}{$\infty$\%} & 0.94 $\pm$ 0.23 & \decline{0.85 $\pm$ 0.36}{10\%} & 0.91 $\pm$ 0.29 & \noimp{0.91 $\pm$ 0.29}{0\%}    \\
\bottomrule
\end{tabular}
}
\end{table}

Table~\ref{tab:all_results} provides a comparative analysis of IR against several other prompting and programmatic baselines: CoT, 2-step CoT, Random, and Greedy, on the GRASP and MiniGrid tasks using the GPT-o3-mini. Importantly, these approaches do not fall under the IPP framework; instead, they operate through natural language prompting and direct action generation without synthesizing executable code. On the GRASP benchmark, IR is the only one that results in a positive average energy score (0.16 $\pm$ 0.21), substantially outperforming CoT (–1.03) and 2-step CoT (–1.91). Programmatic baselines such as Greedy (–3.61) and Random (–4.55) perform even worse. Notably, IR accomplishes this with the lowest API cost among LLM-based methods ($8.0 \times 10^{-4}$). While all reported costs are averaged over 100 instances, it is important to note that IR generates a single final Python program that generalizes to all future instances. As a result, its cost does not grow with the number of instances evaluated. Once the program is synthesized through IR, it can be reused without invoking the model again. In contrast, CoT and 2-step CoT need to generate new action plans for each individual instance, causing their costs to scale linearly with the number of evaluations.

For the three MiniGrid tasks, IR matches or slightly exceeds the other best-performing method. For instance, in the Unlock and Door-Key tasks, IR achieves a perfect or near-perfect reward (0.97 and 0.98 respectively), comparable to or better than CoT and 2-step CoT, but at a substantially lower API cost (as low as $8.1 \times 10^{-4}$ per instance). In the Unlock-Pickup task, IR (0.78 ± 0.34) performs better than CoT (0.48) and nearly matches 2-step CoT (0.84), again at a reduced cost. These results suggest that IR offers a better balance of performance and cost than other LLM-based methods, making it practical for complex planning tasks that demand precise execution.

\begin{table}
\centering
\renewcommand{\arraystretch}{1.5}
\caption{
Comparison of prompting strategies and baseline methods on GRASP and MiniGrid tasks using \textit{GPT-o3-mini}. We evaluate \textbf{Chain-of-Thought (CoT)}, \textbf{2-Step CoT}, and \textbf{Iterative Refinement (IR)}, alongside \textbf{Greedy} and \textbf{Random} programmatic baselines. For GRASP, the metric is \textbf{energy collected}. For MiniGrid, the metric is \textbf{reward} (maximum reward = 1.0). Each cell shows the \textbf{average performance} ± standard deviation, with the \textbf{per-instance API cost} (in USD) in parentheses. Results are averaged over 100 instances per task. \textbf{Bold} values highlight the best-performing strategy per task. \underline{Underlined} values indicate the lowest non-zero cost among methods.
}
\label{tab:all_results}
\resizebox{\textwidth}{!}{%
\begin{tabular}{llccccc}
\toprule
\multirow{2}{*}{\textbf{Benchmark}} & \multirow{2}{*}{\textbf{Task}} 
& \textbf{Random} & \textbf{Greedy} 
& \textbf{CoT} & \textbf{2-step CoT} & \textbf{IR} \\
&& \multicolumn{5}{c}{\textbf{Performance ± Std (Cost)}} \\
\midrule
GRASP & Collect Energy & -4.55 $\pm$ 0.94 (0) & -3.61 $\pm$ 0.26 (0) & -1.03 $\pm$ 0.77 (\$2.0e\textminus3) & -1.91 $\pm$ 1.00 (\$4.0e\textminus3) & \textbf{0.16 $\pm$ 0.21} \underline{(\$8.0e\textminus4)} \\
\addlinespace
\multirow{3}{*}{MiniGrid}
& Unlock         & 0.00 $\pm$ 0.00 (0) & 0.97 $\pm$ 0.01 (0) & 0.97 $\pm$ 0.01 (\$2.0e\textminus3) & 0.97 $\pm$ 0.01 (\$2.0e\textminus3) & 0.97 $\pm$ 0.01 \underline{(\$1.08e\textminus3)} \\
& Door-Key       & 0.00 $\pm$ 0.00 (0) & 0.98 $\pm$ 0.01 (0) & 0.95 $\pm$ 0.17 (\$3.0e\textminus3) & 0.97 $\pm$ 0.10 (\$2.0e\textminus3) & \textbf{0.98 $\pm$ 0.00} \underline{(\$8.4e\textminus4)}\\
& Unlock-Pickup  & 0.00 $\pm$ 0.00 (0) & 0.75 $\pm$ 0.38 (0) & 0.48 $\pm$ 0.47 (\$3.0e\textminus3) & \textbf{0.84 $\pm$ 0.30} (\$3.0e\textminus3) & 0.78 $\pm$ 0.34 \underline{(\$8.1e\textminus4)} \\
\bottomrule
\end{tabular}
}
\end{table}

\begin{figure}
    \centering
    \includegraphics[width=\linewidth]{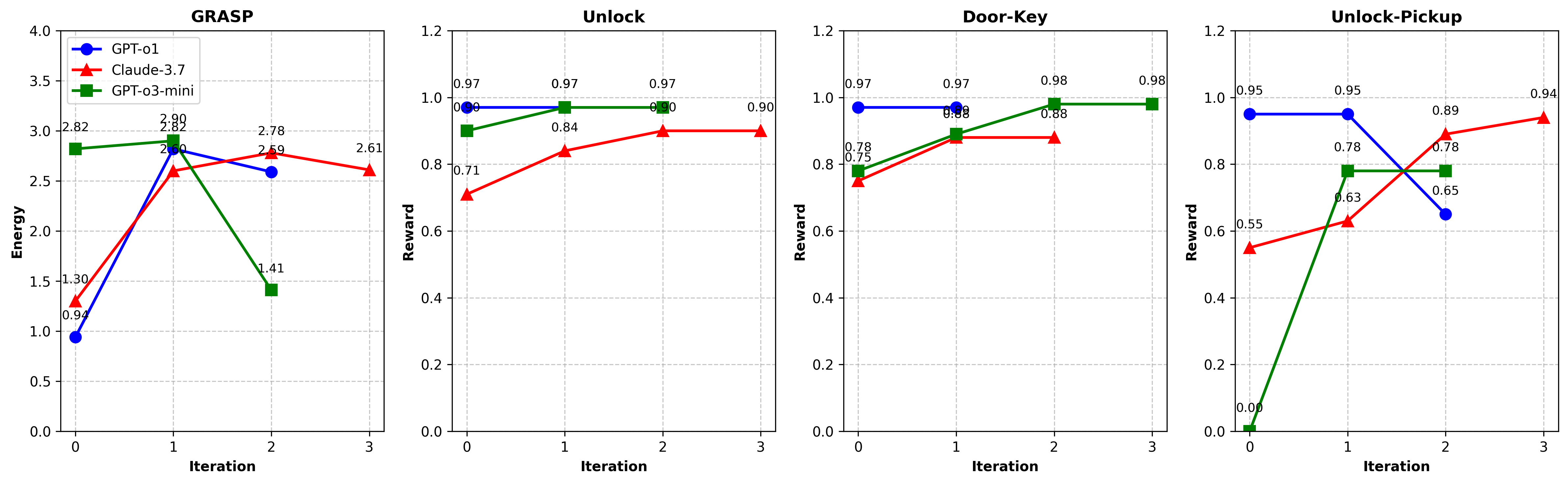}
    \caption{Performance of \textit{GPT-o1}, \textit{Claude-3.7}, and \textit{GPT-o3-mini} across four tasks—\textit{GRASP}, \textit{Unlock}, \textit{Door-Key}, and \textit{Unlock-Pickup}—measured over successive refinement iterations. For GRASP, the metric is energy collected; for MiniGrid tasks, the metric is reward (maximum = 1.0). Iteration 0 corresponds to the initial output produced by the direct generation method.}
    \label{fig:across_iteration}
\end{figure}

Figure~\ref{fig:across_iteration} shows the performance of GPT-o1, Claude-3.7, and GPT-o3-mini using IR across four tasks: GRASP, Unlock, Door Key, and Unlock Pickup. The x-axis shows the number of refinement iterations, with 0 representing direct generation and higher values indicating additional refinement steps. In the GRASP benchmark, GPT-o1 and GPT-o3-mini reach a peak energy score of 2.9 and 2.82 at iteration 1, while Claude-3.7 peaks at 2.78 at iteration 2 before slightly declining.

In the three MiniGrid tasks, GPT-o1 and GPT-o3-mini perform well from the beginning, often achieving near-perfect scores by iteration 1. Claude-3.7 shows the meaningful improvement, particularly in the Unlock and Unlock Pickup tasks, where rewards increase from 0.71 to 0.90 and from 0.55 to 0.94, respectively. Unlike GRASP, performance in MiniGrid tasks almost never declines with additional refinement. Overall, the results show that most performance gains occur within the first one or two iterations of IR. Beyond that point, improvements tend to level off or slightly decrease.

\begin{figure}
    \centering
    \includegraphics[width=0.75\linewidth]{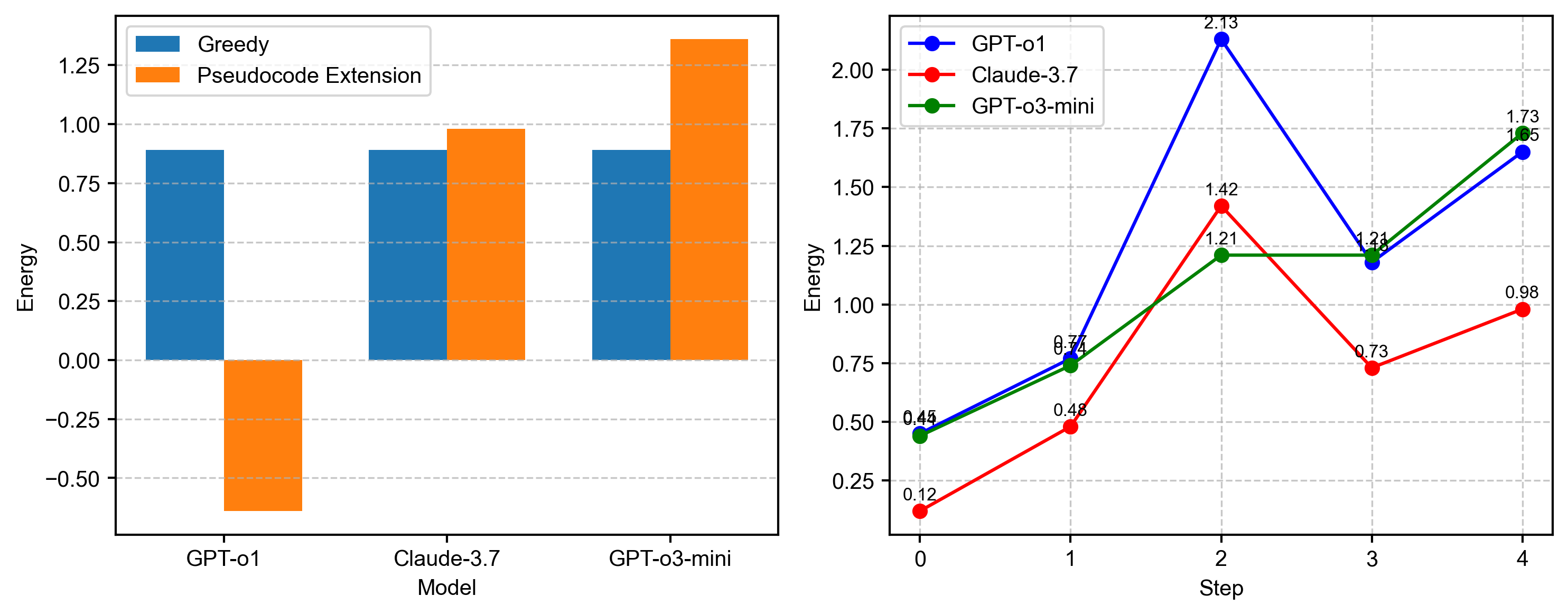}
    \caption{\textbf{Left:} Performance comparison of \textit{Greedy} and \textit{Pseudocode Extension} strategies on the GRASP task, evaluated across \textit{GPT-o1}, \textit{Claude-3.7}, and \textit{GPT-o3-mini}. The results highlight the performance gains enabled by pseudocode-guided reasoning. \textbf{Right:} Step-by-step performance on GRASP, where environment complexity is introduced incrementally over four steps.}
    \label{fig:PE&IC_results}
\end{figure}

Figure~\ref{fig:PE&IC_results} presents two analyses of performance on the GRASP task using GPT-o1, Claude-3.7, and GPT-o3-mini. On the left, it shows the comparison between the Greedy programmatic baseline and the Pseudocode Extension. GPT-o1, which performs reasonably well under the Greedy baseline, fails under Pseudocode Extension with a negative energy score. On the other hand, Claude-3.7 and GPT-o3-mini both benefit from this structured guidance, achieving higher energy scores with Pseudocode Extension, indicating that models with stronger capabilities are better at exploiting high-level pseudocode instructions. On the right, step-by-step performance is shown from step 0 to step 4 when the complexity of the GRASP environment is gradually added. GPT-o1 shows rapid improvement, peaking at step 2 (2.13) before dropping at step 3, then recovering at step 4. Claude-3.7 and GPT-o3-mini show steadier gains, with Claude-3.7 peaking at step 2 and declining at step 3, while GPT-o3-mini shows consistent upward trends. These results suggest that incremental exposure to complexity helps models adapt, though performance can fluctuate depending on how well a model generalizes under new constraints.

\section{Discussion and Limitations}

Our experiments show that LLMs, when guided by a structured iterative process, can generate interpretable and reusable code-based planning policies, even on problems they find initially challenging. The proposed iterative programmatic planning (IPP) framework consistently improves task performance on both GRASP and MiniGrid benchmarks. Even models with a weak initial program benefit significantly from refinement, highlighting the value of execution-based feedback.

Our results demonstrate that both the structure of prompts and the choice of LLMs are critical to planning performance. While direct code generation works in simple scenarios, structured prompting strategies such as pseudocode extension and curriculum-based step-by-step prompting significantly improve outcomes on more complex tasks. For IR, improvements emerge within the first one or two refinement steps, indicating that the process is both effective and computationally efficient. However, LLMs' response to IR varies. GPT-o1,  Claude-3.7, and Gemini-2.5-Pro consistently benefit from our approach, whereas DeepSeek-R1 and GPT-4o sometimes show degraded performance.

A major advantage of IPP is the interpretability of its code-driven policies. Unlike black-box planners or reinforcement learning agents, the generated programs are human-readable and editable, allowing experts to inspect, debug, or adapt plans (see Appendix \ref{app:example_ir_solution} for examples). This transparency is especially valuable in safety-critical domains like robotics. IPP is also more efficient than traditional prompting methods. While chain-of-thought requires inference for each instance, IPP produces a single reusable program, reducing API costs and latency. Rather than competing with reinforcement learning, our aim is to provide a lower-cost, interpretable alternative that can solve similar planning tasks without intensive environment interaction, making it suitable for real-time or resource-constrained applications.

One potential limitation of our framework is that it depends heavily on the language model’s own ability to understand the environment and task rules. While the grid-based setup is fairly simple, models can still make mistakes in tracking the agent’s state, recognizing environment-specific entities, or following task constraints. These mistakes can happen in more complex tasks or when new situations arise that the model hasn’t seen before. Although iterative refinement helps fix some of these problems, there is still no guarantee that the final code will work correctly in all cases. We note, however, that this also applies to RL-based policies.

Another limitation is the cost and time involved in generating and refining the code. The refinement process requires running the model multiple times, each time testing the program on a large number of task instances and asking the model to improve it based on performance. This can take time and compute, especially for harder tasks or when the initial code is far from correct. Also, the way the model uses performance feedback is not very precise. It only sees whether the program did better or worse overall, without knowing exactly what part of the code caused the problem. This can make the refinement less efficient and harder to apply to tasks where small mistakes matter a lot.

\section{Related work}

\textbf{LLMs for Code Generation and Planning} LLMs have demonstrated significant capability in code generation and other code-related tasks, such as bug fixing and program testing \cite{chen2021evaluating, chowdhery2023palm, roziere2023code, wang2023codet5+, jiang2023impact, kang2023large}. This makes LLMs attractive for planning tasks, where structured reasoning, abstraction, and correct execution are critical. Recent frameworks use LLMs to generate and refine plans by combining high-level reasoning with feedback from control or execution. \cite{ajay2023compositional, meng2025audere, wang2024llmˆ}. It has been shown that LLMs can translate natural language instructions into planning formalisms like PDDL \cite{oswald2024large, smirnov2024generating}, or synthesize generalized planners that solve novel tasks via chain-of-thought reasoning and iterative debugging \cite{silver2024generalized}. In embodied AI, LLM has been used to generate code that serves as executable policies for robot control \cite{liang2023code, luo2024large}. This progress suggests the feasibility of using LLMs to produce interpretable and adaptive policies for real-world tasks.

However, much of the prior work has focused on generating action plans for specific instances, often assuming strong task priors. These methods typically emphasize plan generation over policy synthesis, and often lack mechanisms for generalization across varied problem settings. In contrast, our approach treats code as a reusable policy that can adapt to multiple instances within a task. We move beyond static plan generation and instead use LLMs to produce and improve executable programs based on observed performance.

\textbf{Iterative Code Generation} Recent work on iterative code generation LLMs has shown significant progress in code quality and reliability through repeated cycles of feedback and correction \cite{jiang2024training, chen2023teaching}. More specifically, the SELF-REFINE framework showed that LLMs can iteratively improve their outputs by generating feedback and refining their own code, achieving up to 13\% improvement \cite{madaan2023self}. Building on this idea, later approaches automate refinement using external signals such as execution feedback from test suites \cite{ding2024cycle}, compiler diagnostics \cite{bi2024iterative}, and preference-guided comparisons of successful versus failed generations \cite{tao2024codelutra}. Additionally, some research explores multi-agent and interactive refinement, where LLMs act as solvers, verifiers, or critics. This includes self-play for code and test generation \cite{lin2025learning, adnan2025large}, critique-driven updates \cite{zhou2025refinecoder}, and counterexample-guided refinement in inductive synthesis \cite{wei2025codearc}.

While prior work applies iterative methods to typical coding tasks like function completion or bug fixing, our approach extends this paradigm to planning. Instead of refining code for correctness, we use task-level metrics like energy or reward from failures to iteratively guide program generation. Our method improves entire policy programs that can handle a range of environments, not just small code pieces. Unlike traditional tools that use test cases, our framework treats code as a control policy and refines it based on real task performance.

\bibliographystyle{plain}
\bibliography{refs}

\clearpage
\appendix
\section{Appendix}
\addcontentsline{toc}{section}{Appendix}

\subsection*{Contents of the Appendix}
\begin{itemize}
  \item \hyperref[app:experiment]{Additional Experiment Results}
  \item \hyperref[app:example_ir_solution]{Example IPP Solutions}
  \item \hyperref[app:prompts]{Prompts}
  \begin{itemize}
      \item \hyperref[app:grasp_prompt]{GRASP Prompt}
      \item \hyperref[app:unlock_prompt]{MiniGrid - Unlock Prompt}
      \item \hyperref[app:door_key_prompt]{MiniGrid - Door Key Prompt}
      \item \hyperref[app:unlock_pickup_prompt]{MiniGrid - Unlock Pickup Prompt}
  \end{itemize}
  \item \hyperref[app:baseline_code]{Baseline Pseudocodes}
  \begin{itemize}
      \item \hyperref[app:grasp_code]{GRASP}
      \item \hyperref[app:unlock_code]{MiniGrid - Unlock}
      \item \hyperref[app:doorkey_code]{MiniGrid - Door Key}
      \item \hyperref[app:unlockpickup_code]{MiniGrid - Unlock Pickup}
  \end{itemize}
\end{itemize}

\subsection{Additional Experiment Results}\label{app:experiment}

\begin{table}[!htp]
\centering
\renewcommand{\arraystretch}{1.6}
\caption{Performance of LLMs on the GRASP benchmark under different energy distributions (Random, Vertically-skewed, Horizontally-skewed, Cluster, Spiral) with and without obstacles. Two prompting strategies are compared: \textbf{Direct Generation (DG)} and \textbf{Iterative Refinement (IR)}. The numbers represent the average net energy collected; blue/red text indicates the percentage improvement or deterioration of IR over DG. Each cell is mean $\pm$ standard deviation.}
\resizebox{\textwidth}{!}{%
\begin{tabular}{llcccccccccccc}
\toprule
\multirow{2}{*}{\textbf{Model}}  & \multirow{2}{*}{\textbf{Obstacle}} 
& \multicolumn{2}{c}{\textbf{Random}} 
& \multicolumn{2}{c}{\textbf{V-skewed}} 
& \multicolumn{2}{c}{\textbf{H-skewed}} 
& \multicolumn{2}{c}{\textbf{Cluster}} 
& \multicolumn{2}{c}{\textbf{Spiral}} \\
\cmidrule(r){3-4} \cmidrule(r){5-6} \cmidrule(r){7-8} \cmidrule(r){9-10} \cmidrule(r){11-12}
 & 
& DG & IR 
& DG & IR 
& DG & IR 
& DG & IR 
& DG & IR \\
\midrule
\multirow{2}{*}{GPT-4o}         
& With    & -1.40 $\pm$ 3.65 & \improve{-0.38 $\pm$ 3.26}{73\%} & -1.46 $\pm$ 3.63 & \improve{-0.40 $\pm$ 3.27}{73\%} & -1.47 $\pm$ 3.62 & \improve{-0.40 $\pm$ 3.26}{73\%} & -1.93 $\pm$ 3.50 & \improve{-0.62 $\pm$ 3.16}{68\%} & -1.62 $\pm$ 3.51 & \improve{-0.46 $\pm$ 3.08}{72\%} & \\

& Without & -1.32 $\pm$ 3.68 & \improve{-0.36 $\pm$ 3.30}{73\%} & -1.35 $\pm$ 3.71 & \improve{-0.34 $\pm$ 3.32}{75\%} & -1.26 $\pm$ 3.71 & \improve{-0.34 $\pm$ 3.33}{73\%} & -1.96 $\pm$ 3.52 & \improve{-0.69 $\pm$ 3.19}{65\%} & -1.56 $\pm$ 3.54 & \improve{-0.47 $\pm$ 3.14}{70\%} & \\

\addlinespace
\multirow{2}{*}{GPT-o1}         
& With    & 1.09 $\pm$ 2.77 & \improve{2.89 $\pm$ 3.05}{165\%} & 1.04 $\pm$ 2.76 & \improve{2.85 $\pm$ 3.05}{174\%} & 1.01 $\pm$ 2.77 & \improve{2.86 $\pm$ 3.07}{183\%} & 0.50 $\pm$ 2.77 & \improve{2.51 $\pm$ 3.16}{402\%} & 0.81 $\pm$ 2.72 & \improve{2.60 $\pm$ 2.99}{221\%} & \\

& Without & 1.16 $\pm$ 2.79 & \improve{3.04 $\pm$ 3.06}{162\%} & 1.18 $\pm$ 2.78 & \improve{3.03 $\pm$ 3.07}{157\%} & 1.19 $\pm$ 2.79 & \improve{3.10 $\pm$ 3.08}{161\%} & 0.49 $\pm$ 2.75 & \improve{2.59 $\pm$ 3.20}{429\%} & 0.90 $\pm$ 2.75 & \improve{2.76 $\pm$ 3.02}{207\%} & \\

\addlinespace
\multirow{2}{*}{GPT-o3-mini}    
& With    & 2.89 $\pm$ 3.05 & \improve{2.92 $\pm$ 2.86}{1\%} & 2.85 $\pm$ 3.05 & \improve{2.92 $\pm$ 2.88}{2\%} & 2.86 $\pm$ 3.07 & \improve{2.91 $\pm$ 2.87}{2\%} & 2.51 $\pm$ 3.16 & \improve{2.72 $\pm$ 2.80}{8\%} & 2.60 $\pm$ 2.99 & \improve{2.71 $\pm$ 2.75}{4\%} & \\

& Without & 3.04 $\pm$ 3.06 & \noimp{3.04 $\pm$ 2.93}{0\%} & 3.03 $\pm$ 3.07 & \improve{3.05 $\pm$ 2.93}{1\%} & 3.10 $\pm$ 3.08 & \decline{3.07 $\pm$ 2.94}{1\%} & 2.59 $\pm$ 3.20 & \improve{2.78 $\pm$ 2.85}{7\%} & 2.76 $\pm$ 3.02 & \improve{2.85 $\pm$ 2.82}{3\%} & \\

\addlinespace
\multirow{2}{*}{Claude-3.7}     
& With    & 1.32 $\pm$ 3.15 & \improve{2.82 $\pm$ 2.72}{114\%} & 1.19 $\pm$ 3.24 & \improve{2.79 $\pm$ 2.73}{134\%} & 1.15 $\pm$ 3.27 & \improve{2.79 $\pm$ 2.73}{143\%} & 1.30 $\pm$ 3.31 & \improve{2.60 $\pm$ 2.67}{100\%} & 1.33 $\pm$ 3.15 & \improve{2.60 $\pm$ 2.69}{95\%} & \\

& Without & 1.32 $\pm$ 3.17 & \improve{2.92 $\pm$ 2.76}{121\%} & 1.20 $\pm$ 3.26 & \improve{2.91 $\pm$ 2.77}{143\%} & 1.33 $\pm$ 3.25 & \improve{2.96 $\pm$ 2.77}{123\%} & 1.51 $\pm$ 3.35 & \improve{2.66 $\pm$ 2.71}{76\%} & 1.31 $\pm$ 3.18 & \improve{2.76 $\pm$ 2.73}{111\%} & \\

\addlinespace
\multirow{2}{*}{Gemini-2.5-Pro} 
& With    & -0.48 $\pm$ 1.85 & \improve{3.27 $\pm$ 3.05}{781\%} & -0.45 $\pm$ 1.97 & \improve{3.28 $\pm$ 3.06}{829\%} & -0.50 $\pm$ 1.92 & \improve{3.26 $\pm$ 3.06}{752\%} & -0.80 $\pm$ 2.32 & \improve{3.00 $\pm$ 3.00}{475\%} & -0.54 $\pm$ 2.12 & \improve{2.96 $\pm$ 2.94}{648\%} & \\

& Without & -0.44 $\pm$ 1.81 & \improve{3.38 $\pm$ 3.07}{868\%} & -0.41 $\pm$ 1.83 & \improve{3.39 $\pm$ 3.08}{927\%} & -0.38 $\pm$ 1.85 & \improve{3.46 $\pm$ 3.10}{1011\%} & -0.79 $\pm$ 2.36 & \improve{3.09 $\pm$ 3.06}{491\%} & -0.46 $\pm$ 2.09 & \improve{3.12 $\pm$ 2.98}{778\%} & \\

\addlinespace
\multirow{2}{*}{DeepSeek-R1}    
& With    & 1.96 $\pm$ 2.83 & \decline{1.65 $\pm$ 2.21}{16\%} & 1.87 $\pm$ 2.82 & \decline{1.63 $\pm$ 2.17}{13\%} & 1.85 $\pm$ 2.83 & \decline{1.62 $\pm$ 2.16}{12\%} & 1.66 $\pm$ 2.93 & \decline{1.40 $\pm$ 1.88}{16\%} & 1.62 $\pm$ 2.78 & \decline{1.55 $\pm$ 1.98}{4\%} & \\
& Without & 2.00 $\pm$ 2.83 & \decline{1.68 $\pm$ 2.27}{16\%} & 2.02 $\pm$ 2.83 & \decline{1.68 $\pm$ 2.28}{17\%} & 2.06 $\pm$ 2.83 & \decline{1.69 $\pm$ 2.29}{18\%} & 1.73 $\pm$ 2.91 & \decline{1.38 $\pm$ 1.88}{20\%} & 1.78 $\pm$ 2.77 & \decline{1.59 $\pm$ 2.05}{11\%} & \\
\bottomrule
\end{tabular}
}
\end{table}

\begin{table}[!htp]
\centering
\renewcommand{\arraystretch}{1.6}
\caption{Performance of LLMs on the GRASP benchmark under different energy distributions (Random, Vertically-skewed, Horizontally-skewed, Cluster, Spiral) with different cost per step (0 v.s. 0.3). Two prompting strategies are compared: \textbf{Direct Generation (DG)} and \textbf{Iterative Refinement (IR)}. The numbers represent the average net energy collected; blue/red text indicates the percentage improvement or deterioration of IR over DG. Each cell is mean $\pm$ standard deviation.}
\resizebox{\textwidth}{!}{%
\begin{tabular}{llcccccccccccc}
\toprule
\multirow{2}{*}{\textbf{Model}}  & \multirow{2}{*}{\textbf{Cost per step}} 
& \multicolumn{2}{c}{\textbf{Random}} 
& \multicolumn{2}{c}{\textbf{V-skewed}} 
& \multicolumn{2}{c}{\textbf{H-skewed}} 
& \multicolumn{2}{c}{\textbf{Cluster}} 
& \multicolumn{2}{c}{\textbf{Spiral}} \\
\cmidrule(r){3-4} \cmidrule(r){5-6} \cmidrule(r){7-8} \cmidrule(r){9-10} \cmidrule(r){11-12}
 & 
& DG & IR 
& DG & IR 
& DG & IR 
& DG & IR 
& DG & IR \\
\midrule
\multirow{2}{*}{GPT-4o}         
& 0     & 1.64 $\pm$ 2.10 & \improve{2.16 $\pm$ 2.25}{32\%} & 1.60 $\pm$ 2.11 & \improve{2.17 $\pm$ 2.26}{36\%} & 1.63 $\pm$ 2.11 & \improve{2.17 $\pm$ 2.26}{33\%} & 1.06 $\pm$ 1.82 & \improve{1.87 $\pm$ 2.05}{76\%} & 1.41 $\pm$ 1.85 & \improve{2.00 $\pm$ 2.06}{42\%} & \\

& 0.3   & -4.36 $\pm$ 2.11 & \improve{-2.90 $\pm$ 1.91}{33\%} & -4.40 $\pm$ 2.11 & \improve{-2.91 $\pm$ 1.92}{34\%} & -4.36 $\pm$ 2.12 & \improve{-2.91 $\pm$ 1.92}{33\%} & -4.94 $\pm$ 1.82 & \improve{-3.18 $\pm$ 1.81}{36\%} & -4.59 $\pm$ 1.85 & \improve{-2.94 $\pm$ 1.71}{36\%} & \\

\addlinespace
\multirow{2}{*}{GPT-o1}         
& 0     & 3.84 $\pm$ 0.68 & \improve{5.75 $\pm$ 1.36}{50\%} & 3.81 $\pm$ 0.68 & \improve{5.73 $\pm$ 1.35}{50\%} & 3.81 $\pm$ 0.70 & \improve{5.78 $\pm$ 1.38}{52\%} & 3.07 $\pm$ 1.14 & \improve{5.28 $\pm$ 1.79}{72\%} & 3.52 $\pm$ 0.71 & \improve{5.42 $\pm$ 1.33}{54\%} & \\

& 0.3   & -1.59 $\pm$ 0.55 & \improve{0.19 $\pm$ 1.15}{112\%} & -1.59 $\pm$ 0.55 & \improve{0.15 $\pm$ 1.17}{109\%} & -1.61 $\pm$ 0.57 & \improve{0.19 $\pm$ 1.18}{112\%} & -2.08 $\pm$ 0.81 & \improve{-0.18 $\pm$ 1.47}{91\%} & -1.81 $\pm$ 0.55 & \improve{-0.06 $\pm$ 1.10}{97\%} & \\

\addlinespace
\multirow{2}{*}{GPT-o3-mini}    
& 0     & 5.75 $\pm$ 1.36 & \decline{5.68 $\pm$ 1.33}{1\%} & 5.73 $\pm$ 1.35 & \decline{5.69 $\pm$ 1.36}{1\%} & 5.78 $\pm$ 1.38 & \decline{5.70 $\pm$ 1.36}{1\%} & 5.28 $\pm$ 1.79 & \decline{5.25 $\pm$ 1.76}{1\%} & 5.42 $\pm$ 1.33 & \decline{5.38 $\pm$ 1.33}{1\%} & \\

& 0.3   & 0.19 $\pm$ 1.15 & \improve{0.28 $\pm$ 0.67}{47\%} & 0.15 $\pm$ 1.17 & \improve{0.28 $\pm$ 0.67}{87\%} & 0.19 $\pm$ 1.18 & \improve{0.29 $\pm$ 0.66}{53\%} & -0.18 $\pm$ 1.47 & \improve{0.25 $\pm$ 0.62}{239\%} & -0.06 $\pm$ 1.10 & \improve{0.18 $\pm$ 0.50}{400\%} & \\
\addlinespace
\multirow{2}{*}{Claude-3.7}     
& 0     & 4.25 $\pm$ 1.20 & \improve{5.42 $\pm$ 1.12}{28\%} & 4.14 $\pm$ 1.39 & \improve{5.41 $\pm$ 1.13}{31\%} & 4.18 $\pm$ 1.44 & \improve{5.43 $\pm$ 1.16}{30\%} & 4.30 $\pm$ 1.71 & \improve{4.99 $\pm$ 1.60}{16\%} & 4.23 $\pm$ 1.28 & \improve{5.20 $\pm$ 1.17}{23\%} & \\

& 0.3   & -1.61 $\pm$ 1.15 & \improve{0.31 $\pm$ 0.82}{119\%} & -1.75 $\pm$ 1.36 & \improve{0.29 $\pm$ 0.82}{117\%} & -1.70 $\pm$ 1.40 & \improve{0.32 $\pm$ 0.82}{119\%} & -1.48 $\pm$ 1.61 & \improve{0.27 $\pm$ 0.89}{118\%} & -1.59 $\pm$ 1.21 & \improve{0.16 $\pm$ 0.76}{110\%} & \\

\addlinespace
\multirow{2}{*}{Gemini-2.5-Pro} 
& 0     & 1.24 $\pm$ 0.67 & \improve{6.06 $\pm$ 1.54}{389\%} & 1.33 $\pm$ 0.76 & \improve{6.08 $\pm$ 1.55}{357\%} & 1.30 $\pm$ 0.75 & \improve{6.11 $\pm$ 1.59}{370\%} & 1.33 $\pm$ 0.76 & \improve{5.57 $\pm$ 2.03}{319\%} & 1.48 $\pm$ 0.73 & \improve{5.70 $\pm$ 1.53}{285\%} & \\

& 0.3   & -2.16 $\pm$ 0.68 & \improve{0.59 $\pm$ 1.15}{127\%} & -2.19 $\pm$ 0.68 & \improve{0.59 $\pm$ 1.18}{127\%} & -2.18 $\pm$ 0.70 & \improve{0.62 $\pm$ 1.19}{128\%} & -2.93 $\pm$ 1.14 & \improve{0.53 $\pm$ 1.24}{118\%} & -2.47 $\pm$ 0.70 & \improve{0.38 $\pm$ 1.02}{115\%} & \\

\addlinespace
\multirow{2}{*}{DeepSeek-R1}    
& 0     & 4.68 $\pm$ 0.89 & \decline{3.84 $\pm$ 0.68}{18\%} & 4.65 $\pm$ 0.91 & \decline{3.81 $\pm$ 0.68}{18\%} & 4.66 $\pm$ 0.89 & \decline{3.81 $\pm$ 0.70}{18\%} & 4.33 $\pm$ 1.34 & \decline{3.07 $\pm$ 1.14}{29\%} & 4.35 $\pm$ 0.90 & \decline{3.52 $\pm$ 0.71}{19\%} & \\

& 0.3   & -0.72 $\pm$ 0.75 & \improve{-0.51 $\pm$ 0.29}{29\%} & -0.75 $\pm$ 0.77 & \improve{-0.50 $\pm$ 0.30}{33\%} & -0.76 $\pm$ 0.76 & \improve{-0.50 $\pm$ 0.30}{34\%} & -0.95 $\pm$ 1.14 & \improve{-0.29 $\pm$ 0.33}{69\%} & -0.96 $\pm$ 0.74 & \improve{-0.37 $\pm$ 0.26}{61\%} & \\

\bottomrule
\end{tabular}
}
\end{table}

\begin{table}[!htp]
\centering
\caption{Performance of \textit{GPT-o1}, \textit{Claude-3.7}, and \textit{GPT-o3-mini} across four tasks—\textit{GRASP}, \textit{Unlock}, \textit{Door-Key}, and \textit{Unlock-Pickup}—measured over successive refinement iterations. For GRASP, the metric is energy collected; for MiniGrid tasks, the metric is reward (maximum = 1.0). Iteration 0 corresponds to the initial output produced by the direct generation method. Each cell is mean $\pm$ standard deviation.}
\resizebox{\textwidth}{!}{%
\begin{tabular}{llcccc}
\hline
\textbf{Dataset} & \textbf{Model} & \textbf{Iter 0} & \textbf{Iter 1} & \textbf{Iter 2} & \textbf{Iter 3} \\
\hline
\multirow{3}{*}{GRASP (Energy)} 
  & GPT-o1        & 0.94 $\pm$ 2.78 & 2.82 $\pm$ 3.08 & 2.59 $\pm$ 3.34 & -\\
  & Claude-3.7    & 1.30 $\pm$ 3.24 & 2.60 $\pm$ 2.83 & 2.78 $\pm$ 2.73 & 2.61 $\pm$ 2.95\\
  & GPT-o3-mini   & 2.82 $\pm$ 3.08 & 2.90 $\pm$ 2.87 & 1.41 $\pm$ 3.65 & -\\
\hline
\multirow{3}{*}{Unlock (Reward)} 
  & GPT-o1        & 0.97 $\pm$ 0.01 & 0.97 $\pm$ 0.01 & -  & -      \\
  & Claude-3.7    & 0.71 $\pm$ 0.42 & 0.84 $\pm$ 0.33 & 0.90 $\pm$ 0.24 & 0.90 $\pm$ 0.24      \\
  & GPT-o3-mini   & 0.90 $\pm$ 0.24 & 0.97 $\pm$ 0.01 & 0.97 $\pm$ 0.01 & - \\
\hline
\multirow{3}{*}{Door-Key (Reward)} 
  & GPT-o1        &  0.97 $\pm$ 0.00 & 0.97 $\pm$ 0.00 & - & -      \\
  & Claude-3.7    & 0.75 $\pm$ 0.41 & 0.88 $\pm$ 0.28 & 0.88 $\pm$ 0.28 & -      \\
  & GPT-o3-mini   & 0.78 $\pm$ 0.39 & 0.89 $\pm$ 0.27 & 0.98 $\pm$ 0.00 & 0.98 $\pm$ 0.00 \\
\hline
\multirow{3}{*}{Unlock-Pickup (Reward)} 
  & GPT-o1        & 0.95 $\pm$ 0.01 & 0.95 $\pm$ 0.01 & 0.65 $\pm$ 0.18 & - \\
  & Claude-3.7    & 0.55 $\pm$ 0.46 & 0.63 $\pm$ 0.43 & 0.89 $\pm$ 0.21 & 0.94 $\pm$ 0.01 \\
  & GPT-o3-mini   & 0.00 $\pm$ 0.00 & 0.78 $\pm$ 0.34 & 0.78 $\pm$ 0.34 & -\\
\hline
\end{tabular}
}
\end{table}

\begin{table}[!htp]
\centering
\caption{Energy Comparison by Model, Approach, and Step. Left one is for Pseudocode Extension from greedy, right one is for step by step. Only for GRASP. Each cell is mean $\pm$ standard deviation.}
\resizebox{\textwidth}{!}{%
\begin{tabular}{lcc|ccccc}
\hline
\textbf{Model} & \textbf{Greedy} & \textbf{PE} & \textbf{Step 0} & \textbf{Step 1} & \textbf{Step 2} & \textbf{Step 3} & \textbf{Step 4} \\
\hline
GPT-o1        & 0.89 $\pm$ 3.29  & \decline{-0.64 $\pm$ 2.11}{172\%} & 0.45 $\pm$ 3.99 & 0.77 $\pm$ 4.13 & 2.13 $\pm$ 3.67 & 1.18 $\pm$ 1.67 & 1.65 $\pm$ 1.64 \\
Claude-3.7    &   0.89 $\pm$ 3.29    & \improve{0.98 $\pm$ 3.15}{10\%} & 0.12 $\pm$ 3.64 & 0.48 $\pm$ 3.68 & 1.42 $\pm$ 3.54 & 0.73 $\pm$ 1.39 & 0.98 $\pm$ 1.46 \\
GPT-o3-mini   &   0.89 $\pm$ 3.29    & \improve{1.36 $\pm$ 3.79}{52\%} & 0.44 $\pm$ 2.82 & 0.74 $\pm$ 2.88 & 1.21 $\pm$ 3.26 & 1.21 $\pm$ 3.26 & 1.73 $\pm$ 2.91 \\
\hline
\end{tabular}
}
\end{table}

\subsection{Example IPP Solutions}\label{app:example_ir_solution}


\subsubsection{GRASP using Gemini 2.5 Pro}

\textbf{Direct Code Generation}

\begin{scriptsize}
    
\begin{verbatim}
import collections
import math

# --- Helper Functions ---


def is_valid(coord, grid_height, grid_width, grid):
    """Checks if a coordinate is within bounds and not an obstacle."""
    r, c = coord
    return 0 <= r < grid_height and 0 <= c < grid_width and grid[r][c] != "O"


def get_neighbors(coord, grid_height, grid_width, grid, move_directions):
    """Gets valid neighboring coordinates based on move directions."""
    r, c = coord
    potential_moves = []
    # 4 directions
    potential_moves.extend(
        [
            ((r - 1, c), "UP"),
            ((r + 1, c), "DOWN"),
            ((r, c - 1), "LEFT"),
            ((r, c + 1), "RIGHT"),
        ]
    )
    # 8 directions (add diagonals)
    if move_directions == 8:
        potential_moves.extend(
            [
                ((r - 1, c - 1), "UPLEFT"),
                ((r - 1, c + 1), "UPRIGHT"),
                ((r + 1, c - 1), "DOWNLEFT"),
                ((r + 1, c + 1), "DOWNRIGHT"),
            ]
        )

    valid_neighbors = []
    for neighbor_coord, move_name in potential_moves:
        if is_valid(neighbor_coord, grid_height, grid_width, grid):
            valid_neighbors.append((neighbor_coord, move_name))
    return valid_neighbors


def update_position(pos, move):
    """Calculates the new position based on a move action."""
    r, c = pos
    if move == "UP":
        return (r - 1, c)
    if move == "DOWN":
        return (r + 1, c)
    if move == "LEFT":
        return (r, c - 1)
    if move == "RIGHT":
        return (r, c + 1)
    if move == "UPLEFT":
        return (r - 1, c - 1)
    if move == "UPRIGHT":
        return (r - 1, c + 1)
    if move == "DOWNLEFT":
        return (r + 1, c - 1)
    if move == "DOWNRIGHT":
        return (r + 1, c + 1)
    return pos  # Should not happen if move is valid


def find_path_bfs(from_coord, to_coord, grid, move_directions, move_cost):
    """
    Finds the shortest path using Breadth-First Search (BFS).
    Returns (list_of_moves, total_cost) or (None, float('inf')).
    """
    grid_height = len(grid)
    grid_width = len(grid[0])
    queue = collections.deque(
        [(from_coord, [])]
    )  # (coordinate, path_list_of_moves)
    visited = {from_coord}

    while queue:
        current_coord, path = queue.popleft()

        if current_coord == to_coord:
            return path, len(path) * move_cost

        for neighbor_coord, move_name in get_neighbors(
            current_coord, grid_height, grid_width, grid, move_directions
        ):
            if neighbor_coord not in visited:
                visited.add(neighbor_coord)
                new_path = path + [move_name]
                queue.append((neighbor_coord, new_path))

    return None, float("inf")  # No path found


# --- Main Solver Function ---


def solve_energy_game(
    grid, start_pos, max_actions, capacity, move_cost=1, move_directions=4
):
    """
    Solves the energy collection game.

    Args:
        grid (2-D list): The game grid represented as a 2D list of strings.
        max_actions (int): Maximum number of actions allowed.
        capacity (int): Maximum energy tokens the agent can carry.
        move_cost (int): Cost per move action (default 1).
        move_directions (int): 4 or 8 directional movement (default 4).

    Returns:
        list: A list of action strings performed by the agent.
              Returns an empty list if the input grid is invalid.
    """
    # --- Grid Parsing and Initialization ---

    grid_height = len(grid)
    grid_width = len(grid[0]) if grid else 0

    start_pos = None
    initial_energy_coords = set()
    for r in range(grid_height):
        for c in range(grid_width):
            if grid[r][c] == "A":
                start_pos = (r, c)
                grid[r][c] = " "  # Agent location is not a permanent fixture
            elif grid[r][c] == "E":
                initial_energy_coords.add((r, c))

    if start_pos is None:
        print("Error: Agent 'A' not found in the grid.")
        return []

    current_pos = start_pos
    remaining_actions = max_actions
    current_holding = 0
    # Score only counts energy at the start location (initially or dropped)
    score = 1 if (start_pos in initial_energy_coords) else 0
    action_list = []
    # Use a copy of coordinates to track remaining energy
    energy_coords = initial_energy_coords.copy()
    if start_pos in energy_coords:
        energy_coords.remove(
            start_pos
        )  # Don't try to collect from start location initially

    print(
        f"Starting Game: Max Actions={max_actions}, Capacity={capacity}, Start={start_pos}"
    )
    print(f"Initial Energy Count: {len(initial_energy_coords)}")
    # --- Main Loop ---
    while remaining_actions > 0:
        best_trip_plan = None
        max_trip_value = -1  # Using -1 to ensure any valid trip is better

        # --- Evaluate Potential Trips ---
        potential_targets = list(
            energy_coords
        )  # Convert set to list for iteration

        for e_coord in potential_targets:
            # Check capacity *before* calculating paths
            if current_holding >= capacity:
                continue

            # 1. Calculate cost from current position to energy E
            path_to_e, cost_to_e = find_path_bfs(
                current_pos, e_coord, grid, move_directions, move_cost
            )

            if path_to_e is not None:
                # 2. Calculate cost from energy E back to start position
                path_to_start, cost_to_start = find_path_bfs(
                    e_coord, start_pos, grid, move_directions, move_cost
                )

                if path_to_start is not None:
                    actions_take = 1  # Cost of TAKE action
                    actions_drop = 1  # Cost of DROP action
                    num_e_collected = 1  # Evaluating single E trips here
                    total_trip_actions = (
                        cost_to_e + actions_take + cost_to_start + actions_drop
                    )

                    # 3. Check Feasibility (Actions)
                    if total_trip_actions <= remaining_actions:
                        # Simple value: E's collected per action. Higher is better.
                        # Add small epsilon to avoid division by zero if 
                        # total_actions is 0 (shouldn't happen here)
                        trip_value = num_e_collected / (
                            total_trip_actions + 1e-9
                        )

                        # 4. Select Best Feasible Trip
                        if trip_value > max_trip_value:
                            max_trip_value = trip_value
                            best_trip_plan = {
                                "target_e": e_coord,
                                "path_to_target": path_to_e,
                                "cost_to_target": cost_to_e,
                                "path_to_start": path_to_start,
                                "cost_to_start": cost_to_start,
                                "total_actions": total_trip_actions,
                                "collect_count": num_e_collected,
                            }
                            print(
                                f"  Found potential trip to {e_coord}: "\
                                "Cost={total_trip_actions}, Value={trip_value:.2f}"
                            )

        # --- Execute Best Trip or Finish ---
        if best_trip_plan:
            print(
                f"Executing trip to {best_trip_plan['target_e']} "\
                "(Cost: {best_trip_plan['total_actions']},"\
                "Current Actions: {remaining_actions})"
            )
            target_coord = best_trip_plan["target_e"]

            # Move to Target E
            for move in best_trip_plan["path_to_target"]:
                if remaining_actions < move_cost:
                    break  # Cannot afford next move
                action_list.append(move)
                current_pos = update_position(current_pos, move)
                remaining_actions -= move_cost
            if (
                remaining_actions < move_cost and current_pos != target_coord
            ):  # Check if ran out before reaching target
                print("  Ran out of actions before reaching target.")
                break  # Exit main loop if cannot complete action sequence

            # TAKE Action (Check actions first)
            if (
                remaining_actions >= 1
                and current_pos == target_coord
                and target_coord in energy_coords
            ):
                action_list.append("TAKE")
                remaining_actions -= 1
                current_holding += 1
                energy_coords.remove(
                    target_coord
                )  # Remove E from available targets
                print(
                    f"  TAKE at {target_coord}. Holding: {current_holding}. "\
                    "Actions left: {remaining_actions}"
                )

                # Move back to Start
                for move in best_trip_plan["path_to_start"]:
                    if remaining_actions < move_cost:
                        break  # Cannot afford next move
                    action_list.append(move)
                    current_pos = update_position(current_pos, move)
                    remaining_actions -= move_cost
                if (
                    remaining_actions < move_cost and current_pos != start_pos
                ):  # Check if ran out before reaching start
                    print("  Ran out of actions before returning to start.")
                    break  # Exit main loop

                # DROP Action (Check actions and position first)
                if (
                    remaining_actions >= 1
                    and current_pos == start_pos
                    and current_holding > 0
                ):
                    action_list.append("DROP")
                    remaining_actions -= 1
                    score += current_holding  # SCORE UPDATED HERE
                    print(
                        f"  DROP at {start_pos}. Score: {score}. "\
                        "Actions left: {remaining_actions}"
                    )
                    current_holding = 0
                else:
                    # Could happen if ran out of actions exactly when arriving 
                    # at start, or wasn't at start
                    if current_pos != start_pos:
                        print(
                            f"  Warning: Arrived somewhere else ({current_pos}) "\
                            "instead of start after return path."
                        )
                    if current_holding > 0:
                        print(
                            f"  Warning: Failed to DROP {current_holding} energy."
                        )
                    # Continue loop if actions remain, but energy wasn't dropped.
                    # Or break if no actions left
                    if remaining_actions <= 0:
                        break

            else:
                # Failed TAKE (no actions, or not at target, or E already gone somehow)
                if remaining_actions < 1:
                    print(f"  Cannot TAKE at {target_coord}, no actions left.")
                elif current_pos != target_coord:
                    print(
                        f"  Cannot TAKE, not at target {target_coord} (at {current_pos})."
                    )
                elif target_coord not in energy_coords:
                    print(
                        f"  Cannot TAKE at {target_coord}, E is already gone."
                    )
                break  # Critical failure in trip execution

        else:
            # No feasible trips found with remaining actions
            print("No more feasible trips found.")
            break  # Exit main loop

    # --- Final Return if Holding Energy ---
    if (
        current_holding > 0
        and remaining_actions > 0
        and current_pos != start_pos
    ):
        print(
            f"Attempting final return to base with {current_holding} energy "\
            "(Actions left: {remaining_actions})"
        )
        path_home, cost_home = find_path_bfs(
            current_pos, start_pos, grid, move_directions, move_cost
        )

        if (
            path_home is not None and (cost_home + 1) <= remaining_actions
        ):  # +1 for DROP
            print(f"  Executing final return path (Cost: {cost_home + 1})")
            # Execute return path
            for move in path_home:
                if remaining_actions < move_cost:
                    break
                action_list.append(move)
                current_pos = update_position(current_pos, move)
                remaining_actions -= move_cost

            # Final Drop (Check position and actions again)
            if remaining_actions >= 1 and current_pos == start_pos:
                action_list.append("DROP")
                remaining_actions -= 1
                score += current_holding
                print(
                    f"  Final DROP successful. Final Score: {score}. "\
                    "Actions left: {remaining_actions}"
                )
                current_holding = 0
            else:
                print(
                    f"  Final return failed (Actions left: {remaining_actions}, "\
                    "Position: {current_pos}). Energy lost."
                )
                pass
        else:
            print(
                f"  Cannot execute final return (Not enough actions or no path)."
            )
            pass

    elif (
        current_holding > 0
        and current_pos == start_pos
        and remaining_actions >= 1
    ):
        # If ended loop at start but still holding, perform final drop if possible
        print(
            f"Performing final DROP at start (Actions left: {remaining_actions})"
        )
        action_list.append("DROP")
        remaining_actions -= 1
        score += current_holding
        print(
            f"  Final DROP successful. Final Score: {score}. "\
            "Actions left: {remaining_actions}"
        )
        current_holding = 0

    print(f"--- Game Over ---")
    print(f"Final Position: {current_pos}")
    print(f"Final Actions Remaining: {remaining_actions}")
    print(f"Final Energy Held: {current_holding}")
    print(f"Final Score (Energy at {start_pos}): {score}")
    return action_list
\end{verbatim}
\end{scriptsize}

The above shows the result of direct generation by the Gemini-2.5-pro. The code is a simple greedy search that takes round trips around the grid, collecting exactly 1 energy token each time. It selects the energy token for each round trip greedily by picking the one that is closest. But this code confuses movement costs with the total number of actions allowed, which results in reasonable performance when the movement cost is 0, but a negative score when the movement cost is positive, causing the average performance to be negative.

\textbf{Iterative Refinement}

\begin{scriptsize}
    
\begin{verbatim}
import collections
import math

# --- Helper Functions (assumed unchanged, ensure they are present) ---


def is_valid(coord, grid_height, grid_width, grid):
    """Checks if a coordinate is within bounds and not an obstacle."""
    r, c = coord
    return 0 <= r < grid_height and 0 <= c < grid_width and grid[r][c] != "O"


def get_neighbors(coord, grid_height, grid_width, grid, move_directions):
    """Gets valid neighboring coordinates based on move directions."""
    r, c = coord
    potential_moves = []
    # 4 directions
    potential_moves.extend(
        [
            ((r - 1, c), "UP"),
            ((r + 1, c), "DOWN"),
            ((r, c - 1), "LEFT"),
            ((r, c + 1), "RIGHT"),
        ]
    )
    # 8 directions (add diagonals)
    if move_directions == 8:
        potential_moves.extend(
            [
                ((r - 1, c - 1), "UPLEFT"),
                ((r - 1, c + 1), "UPRIGHT"),
                ((r + 1, c - 1), "DOWNLEFT"),
                ((r + 1, c + 1), "DOWNRIGHT"),
            ]
        )

    valid_neighbors = []
    for neighbor_coord, move_name in potential_moves:
        if is_valid(
            neighbor_coord, grid_height, grid_width, grid
        ):  # grid here should be grid_copy for pathfinding
            valid_neighbors.append((neighbor_coord, move_name))
    return valid_neighbors


def update_position(pos, move):
    """Calculates the new position based on a move action."""
    r, c = pos
    if move == "UP":
        return (r - 1, c)
    if move == "DOWN":
        return (r + 1, c)
    if move == "LEFT":
        return (r, c - 1)
    if move == "RIGHT":
        return (r, c + 1)
    if move == "UPLEFT":
        return (r - 1, c - 1)
    if move == "UPRIGHT":
        return (r - 1, c + 1)
    if move == "DOWNLEFT":
        return (r + 1, c - 1)
    if move == "DOWNRIGHT":
        return (r + 1, c + 1)
    return pos


def find_path_bfs(
    from_coord, to_coord, grid_for_bfs, move_directions
):  # Renamed grid to grid_for_bfs for clarity
    """
    Finds the shortest path using Breadth-First Search (BFS).
    Returns (list_of_move_names, num_move_actions) or (None, float('inf')).
    Each move action costs 1.
    """
    grid_height = len(grid_for_bfs)
    grid_width = len(grid_for_bfs[0])

    if from_coord == to_coord:
        return [], 0

    queue = collections.deque([(from_coord, [])])
    visited = {from_coord}

    while queue:
        current_coord, path_moves = queue.popleft()

        for neighbor_coord, move_name in get_neighbors(
            current_coord,
            grid_height,
            grid_width,
            grid_for_bfs,
            move_directions,
        ):
            if neighbor_coord == to_coord:
                return path_moves + [move_name], len(path_moves) + 1

            if neighbor_coord not in visited:
                visited.add(neighbor_coord)
                new_path_moves = path_moves + [move_name]
                queue.append((neighbor_coord, new_path_moves))

    return None, float("inf")


# --- Main Solver Function ---


def solve_grid(
    grid,
    start_pos_param,
    max_actions,
    capacity,
    cost_per_step=1,
    move_directions=4,
):
    if not grid or not grid[0]:
        return []

    grid_height = len(grid)
    grid_width = len(grid[0])

    agent_actual_start_pos = None
    initial_energy_coords = set()
    grid_copy = [
        list(row) for row in grid
    ]  # grid_copy is for BFS pathfinding state

    for r in range(grid_height):
        for c in range(grid_width):
            if (
                grid[r][c] == "A"
            ):  # Check original grid for actual agent position
                if agent_actual_start_pos is None:
                    agent_actual_start_pos = (r, c)
                # else: pass # Multiple 'A's, use first
                grid_copy[r][c] = " "  # Agent location is traversable in copy
            elif grid[r][c] == "E":  # Check original grid for energy
                initial_energy_coords.add((r, c))
                # grid_copy[r][c] remains 'E', is_valid will treat 'E' 
                # as traversable (not 'O')
                # If 'E' should also be ' ' in grid_copy for BFS, change here:
                # grid_copy[r][c] = ' ' # if E cells are treated as empty 
                # for pathfinding once known

    if agent_actual_start_pos is None:
        if (
            0 <= start_pos_param[0] < grid_height
            and 0 <= start_pos_param[1] < grid_width
            and grid[start_pos_param[0]][start_pos_param[1]] != "O"
        ):  # Check original grid for obstacle
            agent_actual_start_pos = start_pos_param
            if (
                grid[agent_actual_start_pos[0]][agent_actual_start_pos[1]]
                == "E"
                and agent_actual_start_pos not in initial_energy_coords
            ):
                initial_energy_coords.add(
                    agent_actual_start_pos
                )  # Ensure E at start_pos_param is known
            grid_copy[agent_actual_start_pos[0]][
                agent_actual_start_pos[1]
            ] = " "
        else:
            return []  # Error

    start_pos = agent_actual_start_pos
    current_pos = start_pos
    remaining_actions = max_actions
    current_holding = 0
    score = 0

    if start_pos in initial_energy_coords:
        score = 1
        initial_energy_coords.remove(start_pos)
        # The location in grid_copy is already ' ' if 'A' was there or made ' ' above.

    action_list = []
    energy_coords_on_grid = initial_energy_coords.copy()

    while remaining_actions > 0:
        if (
            current_holding > 0
            and current_pos == start_pos
            and remaining_actions >= 1
        ):
            action_list.append("DROP")
            remaining_actions -= 1
            score += current_holding
            current_holding = 0
            if remaining_actions <= 0:
                break

        initial_holding_for_this_trip_plan = current_holding

        best_trip_plan = None
        # max_trip_value_efficiency = 0 # OLD
        max_net_gain_for_this_round = (
            0  # NEW: For selecting best trip in current planning phase
        )

        sorted_initial_targets = list(energy_coords_on_grid)
        # MODIFIED: Uncommented and using grid_copy for pathfinding
        sorted_initial_targets.sort(
            key=lambda e_coord: find_path_bfs(
                current_pos, e_coord, grid_copy, move_directions
            )[1]
        )

        for first_e_target_coord in sorted_initial_targets:
            if initial_holding_for_this_trip_plan >= capacity:
                break

            path_to_first_e, actions_to_first_e = find_path_bfs(
                current_pos, first_e_target_coord, grid_copy, move_directions
            )

            if path_to_first_e is None:
                continue

            current_sequence_path_objects = []
            current_sequence_total_actions = 0
            current_sequence_total_move_actions = 0
            current_sequence_energy_collected = 0

            current_sequence_path_objects.append(
                {
                    "moves": path_to_first_e,
                    "actions": actions_to_first_e,
                    "target": first_e_target_coord,
                }
            )
            current_sequence_total_actions += actions_to_first_e + 1
            current_sequence_total_move_actions += actions_to_first_e
            current_sequence_energy_collected += 1

            temp_pos_in_sequence = first_e_target_coord
            temp_available_e_for_sequence = energy_coords_on_grid.copy()
            if (
                first_e_target_coord in temp_available_e_for_sequence
            ):  # Should always be true if logic is correct
                temp_available_e_for_sequence.remove(first_e_target_coord)

            while (
                initial_holding_for_this_trip_plan
                + current_sequence_energy_collected
                < capacity
            ):
                if not temp_available_e_for_sequence:
                    break

                potential_next_extensions = []
                for candidate_e in temp_available_e_for_sequence:
                    path, num_actions_to_cand = find_path_bfs(
                        temp_pos_in_sequence,
                        candidate_e,
                        grid_copy,
                        move_directions,
                    )
                    if path is not None:
                        potential_next_extensions.append(
                            {
                                "coord": candidate_e,
                                "path_moves": path,
                                "actions_to_reach": num_actions_to_cand,
                            }
                        )

                potential_next_extensions.sort(
                    key=lambda x: x["actions_to_reach"]
                )

                added_an_e_this_iteration = False
                for next_e_info in potential_next_extensions:
                    cand_coord = next_e_info["coord"]
                    path_to_cand = next_e_info["path_moves"]
                    actions_to_cand = next_e_info["actions_to_reach"]

                    path_cand_to_start, actions_cand_to_start = find_path_bfs(
                        cand_coord, start_pos, grid_copy, move_directions
                    )

                    if path_cand_to_start is None:
                        continue

                    potential_full_trip_actions = (
                        current_sequence_total_actions
                        + actions_to_cand
                        + 1
                        + actions_cand_to_start
                        + 1
                    )

                    if potential_full_trip_actions <= remaining_actions:
                        current_sequence_path_objects.append(
                            {
                                "moves": path_to_cand,
                                "actions": actions_to_cand,
                                "target": cand_coord,
                            }
                        )
                        current_sequence_total_actions += actions_to_cand + 1
                        current_sequence_total_move_actions += actions_to_cand
                        current_sequence_energy_collected += 1

                        temp_pos_in_sequence = cand_coord
                        if cand_coord in temp_available_e_for_sequence:
                            temp_available_e_for_sequence.remove(cand_coord)
                        added_an_e_this_iteration = True
                        break

                if not added_an_e_this_iteration:
                    break

            final_path_to_start, final_actions_to_start = find_path_bfs(
                temp_pos_in_sequence, start_pos, grid_copy, move_directions
            )

            if final_path_to_start is not None:
                complete_trip_actions = (
                    current_sequence_total_actions + final_actions_to_start + 1
                )
                complete_trip_move_actions = (
                    current_sequence_total_move_actions + final_actions_to_start
                )

                net_energy_gain_this_trip = (
                    current_sequence_energy_collected
                    - (complete_trip_move_actions * cost_per_step)
                )

                if (
                    complete_trip_actions <= remaining_actions
                    and net_energy_gain_this_trip > 0
                ):
                    # MODIFIED: Trip selection logic
                    if net_energy_gain_this_trip > max_net_gain_for_this_round:
                        max_net_gain_for_this_round = net_energy_gain_this_trip
                        best_trip_plan = {
                            "trip_path_segments": current_sequence_path_objects,
                            "return_path_moves": final_path_to_start,
                            "return_path_actions": final_actions_to_start,
                            "total_actions": complete_trip_actions,
                            "num_collected_this_trip": current_sequence_energy_collected,
                            "total_moves_this_trip": complete_trip_move_actions,
                            "net_gain": net_energy_gain_this_trip, \
                            # Store for reference/tie-breaking
                        }
                    elif (
                        net_energy_gain_this_trip == max_net_gain_for_this_round
                    ):  # Tie-breaking
                        # If same net gain, prefer trip with fewer total actions
                        if (
                            best_trip_plan
                            and complete_trip_actions
                            < best_trip_plan["total_actions"]
                        ):
                            best_trip_plan = {
                                "trip_path_segments": current_sequence_path_objects,
                                "return_path_moves": final_path_to_start,
                                "return_path_actions": final_actions_to_start,
                                "total_actions": complete_trip_actions,
                                "num_collected_this_trip": current_sequence_energy_collected,
                                "total_moves_this_trip": complete_trip_move_actions,
                                "net_gain": net_energy_gain_this_trip,
                            }

        if best_trip_plan:
            trip_failed_execution = False
            for i, segment in enumerate(best_trip_plan["trip_path_segments"]):
                target_coord = segment["target"]
                for move in segment["moves"]:
                    if remaining_actions < 1:
                        trip_failed_execution = True
                        break
                    action_list.append(move)
                    current_pos = update_position(current_pos, move)
                    remaining_actions -= 1
                if trip_failed_execution:
                    break

                if (
                    remaining_actions >= 1
                    and current_pos == target_coord
                    and target_coord in energy_coords_on_grid
                ):
                    action_list.append("TAKE")
                    remaining_actions -= 1
                    current_holding += 1
                    energy_coords_on_grid.remove(target_coord)
                else:
                    trip_failed_execution = True
                    break

            if trip_failed_execution:
                continue

            for move in best_trip_plan["return_path_moves"]:
                if remaining_actions < 1:
                    trip_failed_execution = True
                    break
                action_list.append(move)
                current_pos = update_position(current_pos, move)
                remaining_actions -= 1
            if trip_failed_execution:
                continue

            if (
                remaining_actions >= 1
                and current_pos == start_pos
                and current_holding > 0
            ):
                action_list.append("DROP")
                remaining_actions -= 1
                score += current_holding
                current_holding = 0
        else:
            break  # No beneficial trip found

    # Final Return/Drop if Holding Energy
    if current_holding > 0:
        if current_pos == start_pos:
            if remaining_actions >= 1:
                action_list.append("DROP")
                remaining_actions -= 1
                score += current_holding
                current_holding = 0
        else:
            path_home, actions_home = find_path_bfs(
                current_pos, start_pos, grid_copy, move_directions
            )
            if (
                path_home is not None
                and (actions_home + 1) <= remaining_actions
            ):
                for move in path_home:
                    if remaining_actions < 1:
                        break
                    action_list.append(move)
                    current_pos = update_position(current_pos, move)
                    remaining_actions -= 1

                if remaining_actions >= 1 and current_pos == start_pos:
                    action_list.append("DROP")
                    remaining_actions -= 1
                    score += current_holding
                    current_holding = 0

    return action_list
\end{verbatim}
\end{scriptsize}

Here, Gemini-2.5-pro is able to improve the logic in its code by a lot through iterative refinement. In this code, the movement cost and action limit are correctly recognized to be separate constraints and handled properly. In addition to that, the code also improves on the original greedy logic by introducing multi-token round trips, greatly reducing the total number of steps required to collect the same number of tokens. This results in a huge improvement in the final score.

\subsubsection{MiniGrid - Unlock Pickup using Claude 3.7}

\textbf{Direct Code Generation}

\begin{scriptsize}
    
\begin{verbatim}
def solve(grid, start_direction):
    """
    Solve the grid-based game by returning a sequence of actions to pickup the BOX.

    Args:
        grid: 2D list representing the environment
        start_direction: String representing initial direction \
        ("UP", "DOWN", "LEFT", or "RIGHT")

    Returns:
        List of action strings to solve the task
    """
    # Initialize variables
    actions = []
    known_objects = {}
    currently_holding = None

    # Find agent's starting position
    agent_pos = find_agent_position(grid)
    current_pos = agent_pos
    current_direction = start_direction

    # Phase 1: Initial exploration to find KEY and DOOR
    explore_results = explore_current_room(grid, current_pos, current_direction)
    actions.extend(explore_results["actions"])
    current_pos = explore_results["final_pos"]
    current_direction = explore_results["final_direction"]
    update_known_objects(known_objects, explore_results["discovered_objects"])

    # Phase 2: Get KEY
    if "KEY" in known_objects:
        key_pos = known_objects["KEY"]
        nav_results = navigate_to(grid, current_pos, current_direction, key_pos)
        actions.extend(nav_results["actions"])
        current_pos = nav_results["final_pos"]
        current_direction = nav_results["final_direction"]

        # Orient to face the KEY
        orient_results = orient_to_face(current_pos, current_direction, key_pos)
        actions.extend(orient_results["actions"])
        current_direction = orient_results["final_direction"]

        # Pick up KEY
        actions.append("PICKUP")
        currently_holding = "KEY"

    # Phase 3: Go to DOOR and unlock it
    if "DOOR" in known_objects:
        door_pos = known_objects["DOOR"]
        nav_results = navigate_to(
            grid, current_pos, current_direction, door_pos
        )
        actions.extend(nav_results["actions"])
        current_pos = nav_results["final_pos"]
        current_direction = nav_results["final_direction"]

        # Orient to face the DOOR
        orient_results = orient_to_face(
            current_pos, current_direction, door_pos
        )
        actions.extend(orient_results["actions"])
        current_direction = orient_results["final_direction"]

        # Unlock DOOR
        actions.append("UNLOCK")

        # Move through the door
        actions.append("MOVE")
        current_pos = get_position_after_move(current_pos, current_direction)

    # Phase 4: Explore second room to find BOX
    explore_results = explore_current_room(grid, current_pos, current_direction)
    actions.extend(explore_results["actions"])
    current_pos = explore_results["final_pos"]
    current_direction = explore_results["final_direction"]
    update_known_objects(known_objects, explore_results["discovered_objects"])

    # Phase 5: Navigate to BOX
    if "BOX" in known_objects:
        box_pos = known_objects["BOX"]
        nav_results = navigate_to(grid, current_pos, current_direction, box_pos)
        actions.extend(nav_results["actions"])
        current_pos = nav_results["final_pos"]
        current_direction = nav_results["final_direction"]

        # If holding KEY, find empty adjacent cell and drop it
        if currently_holding == "KEY":
            empty_cell_dir = find_empty_adjacent_cell(grid, current_pos)
            if empty_cell_dir:
                # Orient to face empty cell
                orient_results = orient_to_direction(
                    current_direction, empty_cell_dir
                )
                actions.extend(orient_results["actions"])
                current_direction = empty_cell_dir

                # Drop KEY
                actions.append("DROP")
                currently_holding = None

        # Orient to face the BOX
        orient_results = orient_to_face(current_pos, current_direction, box_pos)
        actions.extend(orient_results["actions"])
        current_direction = orient_results["final_direction"]

        # Pick up BOX
        actions.append("PICKUP")

    return actions


def find_agent_position(grid):
    """Find the agent's position in the grid."""
    for i in range(len(grid)):
        for j in range(len(grid[i])):
            if grid[i][j] == "AGENT":
                return (i, j)
    return None


def explore_current_room(grid, start_pos, start_direction):
    """
    Explore the current room using wall-following algorithm.

    Returns:
        Dict containing discovered objects, final position, final direction,
        and actions taken
    """
    actions = []
    discovered_objects = {}
    current_pos = start_pos
    current_direction = start_direction
    visited = set([current_pos])

    # Simple BFS to explore the room
    queue = [(current_pos, current_direction, [])]
    while queue:
        pos, direction, path = queue.pop(0)

        # Check surrounding cells for objects
        for d in ["UP", "RIGHT", "DOWN", "LEFT"]:
            orient_results = orient_to_direction(direction, d)
            new_direction = orient_results["final_direction"]
            new_pos = get_position_after_move(pos, new_direction)

            # Check if position is valid and within room (not a wall)
            if is_valid_position(grid, new_pos):
                cell_content = grid[new_pos[0]][new_pos[1]]
                if (
                    cell_content in ["KEY", "DOOR", "BOX"]
                    and cell_content not in discovered_objects
                ):
                    discovered_objects[cell_content] = new_pos

        # Try moving in all four directions
        for d in ["UP", "RIGHT", "DOWN", "LEFT"]:
            orient_results = orient_to_direction(direction, d)
            new_direction = orient_results["final_direction"]
            new_pos = get_position_after_move(pos, new_direction)

            # Check if position is valid, not visited, and not a wall or door
            if (
                is_valid_position(grid, new_pos)
                and new_pos not in visited
                and grid[new_pos[0]][new_pos[1]] not in ["WALL", "DOOR"]
            ):

                new_path = path.copy()
                new_path.extend(orient_results["actions"])
                new_path.append("MOVE")

                visited.add(new_pos)
                queue.append((new_pos, new_direction, new_path))

        # If we found all objects (KEY, DOOR, BOX), we can stop
        if len(discovered_objects) == 3 or (
            len(discovered_objects) == 2 and "BOX" in discovered_objects
        ):
            break

    # Return the shortest valid path if any
    if queue:
        actions = queue[0][2]
        final_pos = queue[0][0]
        final_direction = queue[0][1]
    else:
        # If no path was found, return original position and direction
        actions = []
        final_pos = start_pos
        final_direction = start_direction

    return {
        "discovered_objects": discovered_objects,
        "final_pos": final_pos,
        "final_direction": final_direction,
        "actions": actions,
    }


def navigate_to(grid, current_pos, current_direction, target_pos):
    """
    Find the shortest path from current_pos to adjacent cell of target_pos.

    Returns:
        Dict containing actions, final position and direction
    """
    # Implementation of A* pathfinding
    from heapq import heappush, heappop

    def heuristic(pos1, pos2):
        # Manhattan distance
        return abs(pos1[0] - pos2[0]) + abs(pos1[1] - pos2[1])

    # Find positions adjacent to target
    adjacent_positions = []
    for d in ["UP", "RIGHT", "DOWN", "LEFT"]:
        adj_pos = get_adjacent_position(target_pos, d)
        if is_valid_position(grid, adj_pos) and grid[adj_pos[0]][
            adj_pos[1]
        ] not in ["WALL", "DOOR", "BOX", "KEY"]:
            adjacent_positions.append(adj_pos)

    if not adjacent_positions:
        return {
            "actions": [],
            "final_pos": current_pos,
            "final_direction": current_direction,
        }

    # Start A* search
    open_set = []
    heappush(open_set, (0, current_pos, current_direction, []))
    closed_set = set()

    while open_set:
        _, pos, direction, path = heappop(open_set)

        if pos in closed_set:
            continue

        closed_set.add(pos)

        # If we've reached an adjacent position to target, return the path
        if pos in adjacent_positions:
            return {
                "actions": path,
                "final_pos": pos,
                "final_direction": direction,
            }

        # Try all four directions
        for d in ["UP", "RIGHT", "DOWN", "LEFT"]:
            orient_results = orient_to_direction(direction, d)
            new_direction = orient_results["final_direction"]
            new_pos = get_position_after_move(pos, new_direction)

            if (
                is_valid_position(grid, new_pos)
                and new_pos not in closed_set
                and grid[new_pos[0]][new_pos[1]]
                not in ["WALL", "DOOR", "BOX", "KEY"]
            ):

                new_path = path.copy()
                new_path.extend(orient_results["actions"])
                new_path.append("MOVE")

                priority = len(new_path) + heuristic(new_pos, target_pos)
                heappush(open_set, (priority, new_pos, new_direction, new_path))

    # If no path is found, return empty
    return {
        "actions": [],
        "final_pos": current_pos,
        "final_direction": current_direction,
    }


def orient_to_face(current_pos, current_direction, target_pos):
    """
    Determine the actions needed to face the target position.

    Returns:
        Dict containing actions and final direction
    """
    # Calculate the direction to face
    row_diff = target_pos[0] - current_pos[0]
    col_diff = target_pos[1] - current_pos[1]

    target_direction = None
    if abs(row_diff) > abs(col_diff):
        # Vertical orientation
        if row_diff < 0:
            target_direction = "UP"
        else:
            target_direction = "DOWN"
    else:
        # Horizontal orientation
        if col_diff < 0:
            target_direction = "LEFT"
        else:
            target_direction = "RIGHT"

    return orient_to_direction(current_direction, target_direction)


def orient_to_direction(current_direction, target_direction):
    """
    Determine the actions needed to change from current to target direction.

    Returns:
        Dict containing actions and final direction
    """
    directions = ["UP", "RIGHT", "DOWN", "LEFT"]
    current_idx = directions.index(current_direction)
    target_idx = directions.index(target_direction)

    # Calculate the number of turns needed
    diff = (target_idx - current_idx) % 4

    actions = []
    if diff == 0:
        # Already facing the target direction
        pass
    elif diff == 1:
        # Turn right once
        actions.append("RIGHT")
    elif diff == 2:
        # Turn right twice or left twice
        actions.extend(["RIGHT", "RIGHT"])
    elif diff == 3:
        # Turn left once
        actions.append("LEFT")

    return {"actions": actions, "final_direction": target_direction}


def find_empty_adjacent_cell(grid, pos):
    """
    Find an empty cell adjacent to the current position.

    Returns:
        Direction to the empty cell or None if not found
    """
    directions = ["UP", "RIGHT", "DOWN", "LEFT"]

    for direction in directions:
        adjacent_pos = get_adjacent_position(pos, direction)
        if (
            is_valid_position(grid, adjacent_pos)
            and grid[adjacent_pos[0]][adjacent_pos[1]] == ""
        ):
            return direction

    return None


def get_position_after_move(pos, direction):
    """Calculate the new position after moving in the given direction."""
    if direction == "UP":
        return (pos[0] - 1, pos[1])
    elif direction == "DOWN":
        return (pos[0] + 1, pos[1])
    elif direction == "LEFT":
        return (pos[0], pos[1] - 1)
    elif direction == "RIGHT":
        return (pos[0], pos[1] + 1)
    return pos


def get_adjacent_position(pos, direction):
    """Get the adjacent position in the given direction without moving."""
    return get_position_after_move(pos, direction)


def is_valid_position(grid, pos):
    """Check if a position is within the grid boundaries."""
    return (
        0 <= pos[0] < len(grid)
        and 0 <= pos[1] < len(grid[0])
        and grid[pos[0]][pos[1]] != "WALL"
    )


def update_known_objects(known_objects, discovered_objects):
    """Update the known_objects dictionary with new discoveries."""
    for obj_type, pos in discovered_objects.items():
        known_objects[obj_type] = pos
\end{verbatim}
\end{scriptsize}

The above shows the code generated by Claude 3.7 with Direct Generation. The model is able to correctly identify the positions of the objects in the correct sequence that is required. But it contains a critical bug where it fails to mark the cell as empty when the agent picks up an object or unlocks the door. This makes it impossible for the agent to reach the box even when the door is unlocked. It is able to succeed in some cases due to another bug related to the `explore\_current\_room` method, which not only identifies the objects in the grid but also makes the agent move to one random object. Therefore, in the cases where the agent moves to the box by chance, it is able to succeed.

\textbf{Iterative Refinement}

\begin{scriptsize}
    
\begin{verbatim}
def solve(grid, start_direction):
    """
    Solve the grid-based game by returning a sequence of actions to pickup the BOX.
    Args:
        grid: 2D list representing the environment
        start_direction: String representing initial direction \
        ("UP", "DOWN", "LEFT", or "RIGHT")
    Returns:
        List of action strings to solve the task
    """
    # Initialize variables
    actions = []
    known_objects = {}
    currently_holding = None

    # Find agent's starting position
    agent_pos = find_agent_position(grid)
    current_pos = agent_pos
    current_direction = start_direction

    # Create a copy of the grid for tracking state changes
    current_grid = [row[:] for row in grid]
    current_grid[agent_pos[0]][
        agent_pos[1]
    ] = ""  # Remove agent from tracking grid

    # Find all objects in grid
    find_all_objects(grid, known_objects)

    # Phase 1: Get KEY
    if "KEY" in known_objects:
        key_pos = known_objects["KEY"]

        # Navigate to a position adjacent to the KEY
        nav_results = navigate_to_adjacent(
            current_grid,
            current_pos,
            current_direction,
            key_pos,
            door_unlocked=False,
        )
        actions.extend(nav_results["actions"])
        current_pos = nav_results["final_pos"]
        current_direction = nav_results["final_direction"]

        # Orient to face the KEY
        orient_results = orient_to_face(current_pos, current_direction, key_pos)
        actions.extend(orient_results["actions"])
        current_direction = orient_results["final_direction"]

        # Pick up KEY
        actions.append("PICKUP")
        currently_holding = "KEY"
        current_grid[key_pos[0]][
            key_pos[1]
        ] = ""  # Remove key from tracking grid

    # Phase 2: Go to DOOR and unlock it
    if "DOOR" in known_objects:
        door_pos = known_objects["DOOR"]

        # Navigate to a position adjacent to the DOOR
        nav_results = navigate_to_adjacent(
            current_grid,
            current_pos,
            current_direction,
            door_pos,
            door_unlocked=False,
        )
        actions.extend(nav_results["actions"])
        current_pos = nav_results["final_pos"]
        current_direction = nav_results["final_direction"]

        # Orient to face the DOOR
        orient_results = orient_to_face(
            current_pos, current_direction, door_pos
        )
        actions.extend(orient_results["actions"])
        current_direction = orient_results["final_direction"]

        # Unlock DOOR
        actions.append("UNLOCK")
        current_grid[door_pos[0]][door_pos[1]] = ""  # Mark door as unlocked

        # Move through the door
        actions.append("MOVE")
        current_pos = get_position_after_move(current_pos, current_direction)

    # Phase 3: Navigate to BOX
    if "BOX" in known_objects:
        box_pos = known_objects["BOX"]

        # Navigate to a position adjacent to the BOX
        nav_results = navigate_to_adjacent(
            current_grid,
            current_pos,
            current_direction,
            box_pos,
            door_unlocked=True,
        )
        actions.extend(nav_results["actions"])
        current_pos = nav_results["final_pos"]
        current_direction = nav_results["final_direction"]

        # If holding KEY, find empty adjacent cell and drop it
        if currently_holding == "KEY":
            # Find an empty cell nearby
            empty_pos = find_empty_cell_near(current_grid, current_pos)
            if empty_pos:
                # Orient to face the empty cell
                orient_results = orient_to_face(
                    current_pos, current_direction, empty_pos
                )
                actions.extend(orient_results["actions"])
                current_direction = orient_results["final_direction"]

                # Drop KEY
                actions.append("DROP")
                currently_holding = None

                # Orient back to BOX
                orient_results = orient_to_face(
                    current_pos, current_direction, box_pos
                )
                actions.extend(orient_results["actions"])
                current_direction = orient_results["final_direction"]

        # Pick up BOX
        actions.append("PICKUP")

    return actions


def find_agent_position(grid):
    """Find the agent's position in the grid."""
    for i in range(len(grid)):
        for j in range(len(grid[i])):
            if grid[i][j] == "AGENT":
                return (i, j)
    return None


def find_all_objects(grid, known_objects):
    """Find all important objects in the grid and add them to known_objects."""
    for i in range(len(grid)):
        for j in range(len(grid[i])):
            if grid[i][j] in ["KEY", "DOOR", "BOX"]:
                known_objects[grid[i][j]] = (i, j)


def navigate_to_adjacent(
    grid, current_pos, current_direction, target_pos, door_unlocked=False
):
    """
    Find a path to a position adjacent to target_pos.
    Returns:
        Dict containing actions, final position and direction
    """
    # If we're already adjacent to the target
    if is_adjacent(current_pos, target_pos):
        return {
            "actions": [],
            "final_pos": current_pos,
            "final_direction": current_direction,
        }

    # Use BFS to find the shortest path
    from collections import deque

    queue = deque(
        [(current_pos, current_direction, [])]
    )  # (pos, direction, actions)
    visited = {(current_pos, current_direction)}

    while queue:
        pos, direction, actions_so_far = queue.popleft()

        # Check if we're adjacent to target
        if is_adjacent(pos, target_pos):
            return {
                "actions": actions_so_far,
                "final_pos": pos,
                "final_direction": direction,
            }

        # Try all possible movements from current position
        for turn_action, new_dir in [
            ([], direction),
            (["RIGHT"], turn_right(direction)),
            (["LEFT"], turn_left(direction)),
            (["RIGHT", "RIGHT"], turn_right(turn_right(direction))),
        ]:
            # Apply turn
            new_actions = actions_so_far.copy()
            new_actions.extend(turn_action)

            # Check if we can move forward
            new_pos = get_position_after_move(pos, new_dir)

            # Check if the move is valid
            is_valid_move = is_valid_position(grid, new_pos) and (
                grid[new_pos[0]][new_pos[1]] == ""
                or (door_unlocked and grid[new_pos[0]][new_pos[1]] == "DOOR")
            )

            # If valid, add move action
            if is_valid_move:
                move_actions = new_actions.copy()
                move_actions.append("MOVE")
                if (new_pos, new_dir) not in visited:
                    visited.add((new_pos, new_dir))
                    queue.append((new_pos, new_dir, move_actions))

    # If no path is found, return empty
    return {
        "actions": [],
        "final_pos": current_pos,
        "final_direction": current_direction,
    }


def find_empty_cell_near(grid, current_pos):
    """Find an empty cell adjacent to the current position."""
    directions = ["UP", "RIGHT", "DOWN", "LEFT"]

    # First check immediate adjacent cells
    for direction in directions:
        new_pos = get_adjacent_position(current_pos, direction)
        if (
            is_valid_position(grid, new_pos)
            and grid[new_pos[0]][new_pos[1]] == ""
        ):
            return new_pos

    # If no immediately adjacent cell, do BFS to find closest empty cell
    from collections import deque

    queue = deque([current_pos])
    visited = {current_pos}

    while queue:
        pos = queue.popleft()

        # Check all adjacent positions
        for direction in directions:
            new_pos = get_adjacent_position(pos, direction)

            if new_pos in visited:
                continue

            if is_valid_position(grid, new_pos):
                if grid[new_pos[0]][new_pos[1]] == "":
                    return new_pos

                if grid[new_pos[0]][new_pos[1]] not in [
                    "WALL",
                    "DOOR",
                    "BOX",
                    "KEY",
                ]:
                    visited.add(new_pos)
                    queue.append(new_pos)

    return None


def is_adjacent(pos1, pos2):
    """Check if two positions are adjacent."""
    return (abs(pos1[0] - pos2[0]) + abs(pos1[1] - pos2[1])) == 1


def orient_to_face(current_pos, current_direction, target_pos):
    """
    Determine the actions needed to face the target position.
    Returns:
        Dict containing actions and final direction
    """
    row_diff = target_pos[0] - current_pos[0]
    col_diff = target_pos[1] - current_pos[1]

    # Determine the direction to face
    if row_diff < 0 and col_diff == 0:
        target_direction = "UP"
    elif row_diff > 0 and col_diff == 0:
        target_direction = "DOWN"
    elif row_diff == 0 and col_diff < 0:
        target_direction = "LEFT"
    elif row_diff == 0 and col_diff > 0:
        target_direction = "RIGHT"
    else:
        # Default to current direction if not clearly horizontally or vertically aligned
        return {"actions": [], "final_direction": current_direction}

    return orient_to_direction(current_direction, target_direction)


def orient_to_direction(current_direction, target_direction):
    """
    Determine the actions needed to change from current to target direction.
    Returns:
        Dict containing actions and final direction
    """
    directions = ["UP", "RIGHT", "DOWN", "LEFT"]
    current_idx = directions.index(current_direction)
    target_idx = directions.index(target_direction)

    # Calculate the number of turns needed
    diff = (target_idx - current_idx) % 4
    actions = []

    if diff == 0:
        # Already facing the target direction
        pass
    elif diff == 1:
        # Turn right once
        actions.append("RIGHT")
    elif diff == 2:
        # Turn right twice
        actions.extend(["RIGHT", "RIGHT"])
    elif diff == 3:
        # Turn left once
        actions.append("LEFT")

    return {"actions": actions, "final_direction": target_direction}


def turn_right(direction):
    """Return the direction after turning right."""
    directions = ["UP", "RIGHT", "DOWN", "LEFT"]
    idx = directions.index(direction)
    return directions[(idx + 1) % 4]


def turn_left(direction):
    """Return the direction after turning left."""
    directions = ["UP", "RIGHT", "DOWN", "LEFT"]
    idx = directions.index(direction)
    return directions[(idx - 1) % 4]


def get_position_after_move(pos, direction):
    """Calculate the new position after moving in the given direction."""
    if direction == "UP":
        return (pos[0] - 1, pos[1])
    elif direction == "DOWN":
        return (pos[0] + 1, pos[1])
    elif direction == "LEFT":
        return (pos[0], pos[1] - 1)
    elif direction == "RIGHT":
        return (pos[0], pos[1] + 1)
    return pos


def get_adjacent_position(pos, direction):
    """Get the adjacent position in the given direction without moving."""
    return get_position_after_move(pos, direction)


def is_valid_position(grid, pos):
    """Check if a position is within the grid boundaries."""
    return (
        0 <= pos[0] < len(grid)
        and 0 <= pos[1] < len(grid[0])
        and grid[pos[0]][pos[1]] != "WALL"
    )
\end{verbatim}

\end{scriptsize}

Through iterative refinement, Claude 3.7 is able to improve its logic and execute the actions in a proper sequence. Unlike in the original attempt, this code sets the cell to an empty state when an object is picked up, enabling correct navigation to the box in every grid instance.

\subsection{Prompts}\label{app:prompts}

\subsubsection{GRASP}\label{app:grasp_prompt}

\textbf{Direct Generation}

1.
\begin{quote}
You are a game playing agent. This game involves a 2-D grid which you have to traverse and get as many energy tokens as possible within a given number of actions (called `max\_actions`) and put the collected energy back in the cell where you started. You are denoted by the letter 'A', the energy tokens are denoted by the letter 'E' and obstacles if any are denoted by the letter 'O'. You can move in '4 ('four') directions: "LEFT", "RIGHT", "UP" and "DOWN"' or 8 ('eight') directions including "UPLEFT", "UPRIGHT", "DOWNLEFT" and "DOWNRIGHT", if diagonals are allowed (boolean parameter called `is\_diagonals\_allowed`). You can also perform two additional actions on the cell you are in currently: "TAKE" (take the energy in the current cell) and "DROP" (drop all the energy tokens you are carrying on to the current cell). You will also have a limit on the amount of energy tokens you can carry at any time (called `carry\_limit`) and also a energy cost per step that you take in any of the different directions (called `cost\_per\_step`) which will be subtracted from the total energy tokens collected at the end.  The goal is to collect as many energy tokens as possible and drop them back in the cell where you started. The energy tokens you are holding onto do not count towards your score. Only the tokens dropped or already present in your starting cell count towards your score. Make sure to drop your collected tokens in your starting cell. Come up with a strategy that would solve this game as optimally as possible. One grid is given as example, you can use that to understand the format. Give me the pseudocode for your strategy. You do not have to focus on writing the actual running code. The output of the algorithm must be the list of actions taken. Make sure to focus on the solution strategy and come with a good strategy that works on any given grid and any given constraints. Ensure your code doesnt cause any errors like out of grid bounds, etc...

\begin{verbatim}
<sample_grid>
    0   1   2   3   4   5   6   7   8   9  10
  +---+---+---+---+---+---+---+---+---+---+---+
0 |   |   | E | E | E | E |   | E | E | E | E |
  +---+---+---+---+---+---+---+---+---+---+---+
1 |   |   | E | E | E |   |   |   |   | E |   |
  +---+---+---+---+---+---+---+---+---+---+---+
2 | E |   |   |   | E |   |   | E | E |   |   |
  +---+---+---+---+---+---+---+---+---+---+---+
3 | E | E | E | E |   | E | E | E |   | E | E |
  +---+---+---+---+---+---+---+---+---+---+---+
4 |   | E |   | A | E |   | E |   | E | E |   |
  +---+---+---+---+---+---+---+---+---+---+---+
5 | E |   | E |   | E | E |   | E |   | E |   |
  +---+---+---+---+---+---+---+---+---+---+---+
6 | E |   | E |   |   |   |   | E | E | E | E |
  +---+---+---+---+---+---+---+---+---+---+---+
7 | E | E |   |   |   |   | E |   | E | E |   |
  +---+---+---+---+---+---+---+---+---+---+---+
8 | E |   | E | E | E |   |   |   | E | E |   |
  +---+---+---+---+---+---+---+---+---+---+---+
9 |   |   | E | E | E |   | E |   |   |   |   |
  +---+---+---+---+---+---+---+---+---+---+---+
10|   | E | E |   | E | E |   | E |   | E | E |
  +---+---+---+---+---+---+---+---+---+---+---+
</sample_grid>
\end{verbatim}
\end{quote}
2.
\begin{quote}
Now, using your pseudocode, write Python code for this task.
Provide a function `solve` that can be called with inputs `grid, start\_pos, carry\_limit, cost\_per\_step, is\_diagonals\_allowed, max\_actions` to get the list of actions.
\end{quote}

\textbf{Iterative Refinement}
\begin{quote}
You were tasked with writing code to play a game optimally. But your code didn’t perform well on certain types of grids. You are now given the entire problem statement, along with your code and some grid examples where your code doesnt perform properly. Modify your code to perform well on these types of grids.  \\
<game\_instructions>  \\
You are a game playing agent. This game involves a 2-D grid which you have to traverse and get as many energy tokens as possible within a given number of actions (called `max\_actions`) and put the collected energy back in the cell where you started. You are denoted by the letter 'A', the energy tokens are denoted by the letter 'E' and obstacles if any are denoted by the letter 'O'. You can move in '4 ('four') directions: "LEFT", "RIGHT", "UP" and "DOWN"' or 8 ('eight') directions including "UPLEFT", "UPRIGHT", "DOWNLEFT" and "DOWNRIGHT", if diagonals are allowed (boolean parameter called `is\_diagonals\_allowed`). You can also perform two additional actions on the cell you are in currently: "TAKE" (take the energy in the current cell) and "DROP" (drop all the energy tokens you are carrying on to the current cell). You will also have a limit on the amount of energy tokens you can carry at any time (called `carry\_limit`) and also a energy cost per step that you take in any of the different directions (called `cost\_per\_step`) which will be subtracted from the total energy tokens collected at the end. The goal is to collect as many energy tokens as possible and drop them back in the cell where you started. The energy tokens you are holding onto do not count towards your score. Only the tokens dropped or already present in your starting cell count towards your score. Make sure to drop your collected tokens in your starting cell. Ensure your code doesnt cause any errors like out of grid bounds, etc...  \\    
</game\_instructions>  \\
<your\_code>\\
\# The code from the previous iteration inserted here\\
</your\_code>\\
<sample\_grids>    \\
\# Grid 1 with worst code performance inserted here\\
---------\\
\# Grid 2 with worst code performance inserted here\\
---------\\
\# Grid 3 with worst code performance inserted here\\
</sample\_grids>
\end{quote}

\textbf{Pseudocode Extension from Greedy}

1.
\begin{quote}
You are a game playing agent. This game involves a 2-D grid which you have to traverse and get as many energy tokens as possible within a given number of actions (called `max\_actions`) and put the collected energy back in the cell where you started. You are denoted by the letter 'A', the energy tokens are denoted by the letter 'E' and obstacles if any are denoted by the letter 'O'. You can move in '4 ('four') directions: "LEFT", "RIGHT", "UP" and "DOWN"' or 8 ('eight') directions including "UPLEFT", "UPRIGHT", "DOWNLEFT" and "DOWNRIGHT", if diagonals are allowed (boolean parameter called `is\_diagonals\_allowed`). You can also perform two additional actions on the cell you are in currently: "TAKE" (take the energy in the current cell) and "DROP" (drop all the energy tokens you are carrying on to the current cell). You will also have a limit on the amount of energy tokens you can carry at any time (called `carry\_limit`) and also a energy cost per step that you take in any of the different directions (called `cost\_per\_step`) which will be subtracted from the total energy tokens collected at the end. The goal is to collect as many energy tokens as possible and drop them back in the cell where you started. The energy tokens you are holding onto do not count towards your score. Only the tokens dropped or already present in your starting cell count towards your score. Make sure to drop your collected tokens in your starting cell. Come up with a strategy that would solve this game as optimally as possible. One grid is given as example, you can use that to understand the format. Give me the pseudocode for your strategy. You do not have to focus on writing the actual running code. The output of the algorithm must be the list of actions taken. Make sure to focus on the solution strategy and come with a good strategy that works on any given grid and any given constraints. Ensure your code doesnt cause any errors like out of grid bounds, etc...  
You are also given the pseudocode of a sample greedy algorithm for this task. You can use this directly or extend it to improve it further. This code is purely given as a reference for the solution logic. You should follow your own coding style / semantics in your pseudocode.  

\begin{verbatim}
<sample_grid>
    0   1   2   3   4   5   6   7   8   9  10
  +---+---+---+---+---+---+---+---+---+---+---+
0 |   |   | E | E | E | E |   | E | E | E | E |
  +---+---+---+---+---+---+---+---+---+---+---+
1 |   |   | E | E | E |   |   |   |   | E |   |
  +---+---+---+---+---+---+---+---+---+---+---+
2 | E |   |   |   | E |   |   | E | E |   |   |
  +---+---+---+---+---+---+---+---+---+---+---+
3 | E | E | E | E |   | E | E | E |   | E | E |
  +---+---+---+---+---+---+---+---+---+---+---+
4 |   | E |   | A | E |   | E |   | E | E |   |
  +---+---+---+---+---+---+---+---+---+---+---+
5 | E |   | E |   | E | E |   | E |   | E |   |
  +---+---+---+---+---+---+---+---+---+---+---+
6 | E |   | E |   |   |   |   | E | E | E | E |
  +---+---+---+---+---+---+---+---+---+---+---+
7 | E | E |   |   |   |   | E |   | E | E |   |
  +---+---+---+---+---+---+---+---+---+---+---+
8 | E |   | E | E | E |   |   |   | E | E |   |
  +---+---+---+---+---+---+---+---+---+---+---+
9 |   |   | E | E | E |   | E |   |   |   |   |
  +---+---+---+---+---+---+---+---+---+---+---+
10|   | E | E |   | E | E |   | E |   | E | E |
  +---+---+---+---+---+---+---+---+---+---+---+
</sample_grid>
\end{verbatim}
  
<example\_greedy\_pseudocode>  

\begin{scriptsize}
\begin{verbatim}
    
```  
# This greedy algorithm doesn't include considerations for cost_per_step and carry_limit  
FUNCTION GREEDY_SOLVE(grid, start_position, is_diagonals_allowed, max_actions):
    actions = []
    current_position = start_position
    actions_remaining = max_actions
    while actions_remaining > 0:
        # Perform Breadth First Search (BFS) to find the nearest energy token
        nearest_energy_position, path = find_nearest_energy_bfs(
            grid, current_position, is_diagonals_allowed
        )
        if nearest_energy_position is None:
            # No more energy tokens available in the grid
            break
        path_length = len(path)
        # Find the number of steps required to return to the start
        # by backtracking in the same path to reach here
        steps_needed_to_return = len(
            [action for action in actions if action not in ['TAKE', 'DROP']]
        )
        # If there is not enough steps to go to the nearest energy token
        # and come back to the start, return to start and end the trip
        # The +2 is for the TAKE and DROP to be performed
        if 2*path_length + steps_needed_to_return + 2 > actions_remaining:
            # Actions to backtrack the same path taken to reach here
            return_actions = return_to_start(actions)
            actions.extend(return_actions)
            actions.append('DROP')
            return actions
        # The remaining steps allow going to the nearest energy token before
        # going back to start
        for direction in path:
            if actions_remaining <= 0:
                break
            move_agent(direction)
            actions.append(direction)
            actions_remaining -= 1
        actions.append('TAKE')
        actions_remaining -= 1
    return actions  
```  
\end{verbatim}
\end{scriptsize}

</example\_greedy\_pseudocode>
\end{quote}

2.
\begin{quote}
Now, using your pseudocode, write Python code for this task.
Provide a function `solve` that can be called with inputs `grid, start\_pos, carry\_limit, cost\_per\_step, is\_diagonals\_allowed, max\_actions` to get the list of actions.
\end{quote}

\textbf{Step by Step}

1.
\begin{quote}
You are a game playing agent. This game involves a 2-D grid which you have to traverse and get as many energy tokens as possible within a given number of actions (called `max\_actions`) and put the collected energy back in the cell where you started. You are denoted by the letter 'A', the energy tokens are denoted by the letter 'E'. You can move in 4 directions: "LEFT", "RIGHT", "UP" and "DOWN"'. You can also perform two additional actions on the cell you are in currently: "TAKE" (take the energy in the current cell) and "DROP" (drop all the energy tokens you are carrying on to the current cell). The goal is to collect as many energy tokens as possible and drop them back in the cell where you started. The energy tokens you are holding onto do not count towards your score. Only the tokens dropped or already present in your starting cell count towards your score. Make sure to drop your collected tokens in your starting cell. Come up with a strategy that would solve this game as optimally as possible. One grid is given as example, you can use that to understand the format. Give me the pseudocode for your strategy. You do not have to focus on writing the actual running code. The output of the algorithm must be the list of actions taken. Make sure to focus on the solution strategy and come with a good strategy that works on any given grid and any given constraints. Ensure your code doesnt cause any errors like out of grid bounds, etc...  
\begin{verbatim}
<sample_grid>
    0   1   2   3   4   5   6   7   8   9  10
  +---+---+---+---+---+---+---+---+---+---+---+
0 |   |   | E | E | E | E |   | E | E | E | E |
  +---+---+---+---+---+---+---+---+---+---+---+
1 |   |   | E | E | E |   |   |   |   | E |   |
  +---+---+---+---+---+---+---+---+---+---+---+
2 | E |   |   |   | E |   |   | E | E |   |   |
  +---+---+---+---+---+---+---+---+---+---+---+
3 | E | E | E | E |   | E | E | E |   | E | E |
  +---+---+---+---+---+---+---+---+---+---+---+
4 |   | E |   | A | E |   | E |   | E | E |   |
  +---+---+---+---+---+---+---+---+---+---+---+
5 | E |   | E |   | E | E |   | E |   | E |   |
  +---+---+---+---+---+---+---+---+---+---+---+
6 | E |   | E |   |   |   |   | E | E | E | E |
  +---+---+---+---+---+---+---+---+---+---+---+
7 | E | E |   |   |   |   | E |   | E | E |   |
  +---+---+---+---+---+---+---+---+---+---+---+
8 | E |   | E | E | E |   |   |   | E | E |   |
  +---+---+---+---+---+---+---+---+---+---+---+
9 |   |   | E | E | E |   | E |   |   |   |   |
  +---+---+---+---+---+---+---+---+---+---+---+
10|   | E | E |   | E | E |   | E |   | E | E |
  +---+---+---+---+---+---+---+---+---+---+---+
</sample_grid>
\end{verbatim}
\end{quote}
2.
\begin{quote}
Now, using your pseudocode, write Python code for this task.
Provide a function `solve` that can be called with inputs `grid, start\_position, max\_actions` to get the list of actions.
\end{quote}
3.
\begin{quote}
Now, you are allowed to move in 8 directions, "UPLEFT", "UPRIGHT", "DOWNLEFT", "DOWNRIGHT" in addition to the 4 initial movement directions. You will be given a boolean parameter `is\_diagonals\_allowed` to indicate whether you are allowed to use the new 4 diagonal movement directions.
Update your code accordingly.
\end{quote}
4.
\begin{quote}
Now, there is an additional constraint: There is a limit on the number of energy tokens you can carry at any time. You will be given a integer parameter `carry\_limit` which represents the maximum number of energy tokens you can carry at any time.
Update your code accordingly.
\end{quote}
5.
\begin{quote}
Now, there is an additional constraint: There is a cost associated with a step in any direction. You will be given a float parameter `cost\_per\_step` which represents the cost associated with a step in any direction which will be subtracted at the end from the total number of energy tokens you collect.
Update your code accordingly.
\end{quote}
6.
\begin{quote}
Now, there is an additional constraint: There can be obstacles in the grid. They will be represented by the letter 'O' in the grid. You cannot stand on or go through a cell with an obstacle.
Update your code accordingly.
\end{quote}

\textbf{Direct Answer}

\begin{quote}
You are a game playing agent. This game involves a 2-D grid which you have to traverse and get as many energy tokens as possible within 20 actions and put the collected energy back in the cell where you started. Positions in the grid will be in the format [row, column] and the numbering starts from zero. [0, 0] denotes top left corner. You are denoted by the letter 'A', the energy tokens are denoted by the letter 'E' and obstacles if any are denoted by the letter 'O'. You can move in 8 directions: "LEFT", "RIGHT", "UP", "DOWN", "UPLEFT", "UPRIGHT", "DOWNLEFT", "DOWNRIGHT". You can also perform two additional actions on the cell you are in currently: "TAKE" (take the energy in the current cell) and "DROP" (drop the energy you are carrying on to the current cell). At any point you can carry only 2 energy tokens. Each step costs you 0.3 energy tokens. Your task is to collect as many energy tokens as possible and drop them back in the cell where you started. The energy tokens you are holding onto do not count towards your score. Only the tokens dropped or already present in your starting cell count towards your score.
    
And finally when you are ready to give your final answer, you can output in this form:
<final\_answer>
    {["RIGHT", "TAKE", ..., "DROP"]}
</final\_answer>

Your final answer must be a list of actions where each action is one of the following:\\
- a direction to move in ("LEFT", "RIGHT", "UP", "DOWN", "UPLEFT", "UPRIGHT", "DOWNLEFT", "DOWNRIGHT")\\
- "TAKE" to take the energy token\\
- "DROP" to drop the energy token\\

Make sure to check the following in your final answer:\\
- You must not move out of the grid.\\
- You must not take energy from a cell that does not have energy.\\
- You must not use more than 20 actions (includes movement, take and drop).\\
- You must not use any other actions than the ones specified above.\\
- After all the actions, only the tokens dropped or already present in your starting cell count towards your score. Tokens you are holding onto do not count. \\
- At any point you can carry only 2 energy tokens\\
- Each step costs you 0.3 energy tokens\\
Adhering to these rules is VERY IMPORTANT.\\

Think about this before writing your output. Use only xml tags for formatting your output. Do not use json.

<grid>\\
\# Insert the input grid here\\
</grid>\\
\end{quote}

\textbf{CoT}

\begin{quote}
You are a game playing agent. This game involves a 2-D grid which you have to traverse and get as many energy tokens as possible within 20 actions and put the collected energy back in the cell where you started. Positions in the grid will be in the format [row, column] and the numbering starts from zero. [0, 0] denotes top left corner. You are denoted by the letter 'A', the energy tokens are denoted by the letter 'E' and obstacles if any are denoted by the letter 'O'. You can move in 8 directions: "LEFT", "RIGHT", "UP", "DOWN", "UPLEFT", "UPRIGHT", "DOWNLEFT", "DOWNRIGHT". You can also perform two additional actions on the cell you are in currently: "TAKE" (take the energy in the current cell) and "DROP" (drop the energy you are carrying on to the current cell). At any point you can carry only 2 energy tokens. Each step costs you 0.3 energy tokens. Your task is to collect as many energy tokens as possible and drop them back in the cell where you started. The energy tokens you are holding onto do not count towards your score. Only the tokens dropped or already present in your starting cell count towards your score.
    
And finally when you are ready to give your final answer, you can output in this form:
<final\_answer>
    {["RIGHT", "TAKE", ..., "DROP"]}
</final\_answer>

Your final answer must be a list of actions where each action is one of the following:\\
- a direction to move in ("LEFT", "RIGHT", "UP", "DOWN", "UPLEFT", "UPRIGHT", "DOWNLEFT", "DOWNRIGHT")\\
- "TAKE" to take the energy token\\
- "DROP" to drop the energy token\\

Make sure to check the following in your final answer:\\
- You must not move out of the grid.\\
- You must not take energy from a cell that does not have energy.\\
- You must not use more than 20 actions (includes movement, take and drop).\\
- You must not use any other actions than the ones specified above.\\
- After all the actions, only the tokens dropped or already present in your starting cell count towards your score. Tokens you are holding onto do not count. \\
- At any point you can carry only 2 energy tokens\\
- Each step costs you 0.3 energy tokens\\
Adhering to these rules is VERY IMPORTANT.\\

Think about this before writing your output. Use only xml tags for formatting your output. Do not use json.

<grid>\\
\# Insert the input grid here\\
</grid>\\

Let's think step by step.
\end{quote}

\textbf{2-Step CoT}

1.
\begin{quote}
You are a game playing agent. This game involves a 2-D grid which you have to traverse and get as many energy tokens as possible within 20 actions and put the collected energy back in the cell where you started. Positions in the grid will be in the format [row, column] and the numbering starts from zero. [0, 0] denotes top left corner. You are denoted by the letter 'A', the energy tokens are denoted by the letter 'E' and obstacles if any are denoted by the letter 'O'. You can move in 8 directions: "LEFT", "RIGHT", "UP", "DOWN", "UPLEFT", "UPRIGHT", "DOWNLEFT", "DOWNRIGHT". You can also perform two additional actions on the cell you are in currently: "TAKE" (take the energy in the current cell) and "DROP" (drop the energy you are carrying on to the current cell). At any point you can carry only 2 energy tokens. Each step costs you 0.3 energy tokens. Your task is to collect as many energy tokens as possible and drop them back in the cell where you started. The energy tokens you are holding onto do not count towards your score. Only the tokens dropped or already present in your starting cell count towards your score.
    
And finally when you are ready to give your final answer, you can output in this form:
<final\_answer>
    {["RIGHT", "TAKE", ..., "DROP"]}
</final\_answer>

Your final answer must be a list of actions where each action is one of the following:\\
- a direction to move in ("LEFT", "RIGHT", "UP", "DOWN", "UPLEFT", "UPRIGHT", "DOWNLEFT", "DOWNRIGHT")\\
- "TAKE" to take the energy token\\
- "DROP" to drop the energy token\\

Make sure to check the following in your final answer:\\
- You must not move out of the grid.\\
- You must not take energy from a cell that does not have energy.\\
- You must not use more than 20 actions (includes movement, take and drop).\\
- You must not use any other actions than the ones specified above.\\
- After all the actions, only the tokens dropped or already present in your starting cell count towards your score. Tokens you are holding onto do not count. \\
- At any point you can carry only 2 energy tokens\\
- Each step costs you 0.3 energy tokens\\
Adhering to these rules is VERY IMPORTANT.\\

Think about this before writing your output. Use only xml tags for formatting your output. Do not use json.

You should first give your first 10 actions. Based on those actions, an updated grid will be given to you, after which you will have to give your entire action sequence.

<grid>\\
\# Insert the input grid here\\
</grid>\\
\end{quote}

2.
\begin{quote}
The updated grid is:\\
<updated\_grid>\\
{grid\_str}\\
</updated\_grid>\\
So far, you have collected [\# Number of energy tokens collected so far] energy tokens and you have incurred a cost of [\# Cost of actions so far] for using [\# Number of actions so far] moves (0.3 each). Now complete your actions and give your entire actions sequence.
\end{quote}

\subsubsection{MiniGrid - Unlock}\label{app:unlock_prompt}

\textbf{Direct Generation}

1.
\begin{quote}
    You are a game-playing agent in a 2D grid-based environment. Your objective is to unlock a `DOOR`.

\#\#\# Your Task:
You are **not** expected to solve the game for a specific instance.
Instead, your task is to **write pseudocode** that outlines a general strategy for solving any valid instance of this game.  
Your pseudocode should describe an algorithm that works on **any grid** and under **any valid configuration** of the objects and the agent’s starting direction. The score will be based on the number of actions taken, so the fewer actions you take, the better your score.

\#\#\# Environment Description:
- The grid is represented as a 2D array, with the top-left corner at `[0][0]`. Each cell can contain one of the following objects:
  - `WALL`, `DOOR`, `KEY`, `AGENT` (You), or an empty string (`""`).
- You start in a specific cell facing one of four directions: `UP`, `DOWN`, `LEFT`, or `RIGHT` (this will be provided).
- Two rooms are separated by `WALL`s, with a single `DOOR` connecting them.

\#\#\# Actions:
You can perform the following actions:
- `LEFT`: Turn 90° to the left.
- `RIGHT`: Turn 90° to the right.
- `MOVE`: Move one step forward in the direction you're facing.
- `PICKUP`: Pick up an object located in the cell *directly in front* of you.
- `DROP`: Drop the object you're holding into the cell *directly in front* of you.
- `UNLOCK`: Unlock the door located *directly in front* of you, if you are holding the `KEY`.

\#\#\# Rules:
- Before picking up a new object, you must `DROP` the one you’re holding into an **empty adjacent cell** (no `WALL`, `DOOR`, or object).
- You can only `PICKUP` or `UNLOCK` if:
  - The target cell is directly in front of you (i.e., one step in the direction you're facing).
  - You are adjacent to that cell.
- You cannot enter a cell that contains any object unless it’s been picked up.

\#\#\# Input:
You will be provided:
- The complete 2D grid as a list of lists (example below).
- Your starting facing direction.

Below is a sample input grid and starting direction along with a sample solution to this grid.
\begin{scriptsize}

\begin{verbatim}
<sample_grid>
[
["WALL","WALL","WALL","WALL","WALL","WALL","WALL","WALL","WALL","WALL","WALL"],
["WALL","","","","KEY","WALL","","","","","WALL"],
["WALL","","","AGENT","","WALL","","","","","WALL"],
["WALL","","","","","DOOR","","","","","WALL"],
["WALL","","","","","WALL","","","","","WALL"],
["WALL","WALL","WALL","WALL","WALL","WALL","WALL","WALL","WALL","WALL","WALL"]
]
</sample_grid>
<start_direction>
RIGHT
</start_direction>
<sample_solution>
[
  # Starting at (2, 3) facing RIGHT
  "LEFT", "MOVE", "RIGHT" # Move to (1, 3) and face RIGHT to face the KEY at (1, 4)
  "PICKUP", # Pick up the KEY at (1, 4)
  "RIGHT", "MOVE", "MOVE", "LEFT", # Move to (3, 4) and face RIGHT to face the DOOR\
  at (3, 5)
  "UNLOCK", # Unlock the DOOR at (3, 5)
]
<sample_solution>
\end{verbatim}
\end{scriptsize}
\end{quote}

2.
\begin{quote}
    Now, using the pseudocode you previously wrote, translate it into Python code.

Write a function `solve(grid, start\_direction)` that returns the list of actions to complete the task.  \\
- `grid` is a 2D list representing the environment.\\
- `start\_direction` is a string: one of `"UP"`, `"DOWN"`, `"LEFT"`, or `"RIGHT"`.
\end{quote}

\textbf{Iterative Refinement}

\begin{quote}
You were tasked with writing code to play a game optimally. But your code didn’t perform well on certain types of grids. You are now given the entire problem statement, along with your code and some grid examples where your code doesnt perform properly. Observe and understand from the sample grids given, where your code goes wrong. Modify your code to perform well on these types of grids.  

<game\_instructions> \\ 
You are a game-playing agent in a 2D grid-based environment. Your objective is to unlock a `DOOR`.

\#\#\# Environment Description:\\
- The grid is represented as a 2D array, with the top-left corner at `[0][0]`. Each cell can contain one of the following objects:\\
  - `WALL`, `DOOR`, `KEY`, `AGENT` (You), or an empty string (`""`).\\
- You start in a specific cell facing one of four directions: `UP`, `DOWN`, `LEFT`, or `RIGHT` (this will be provided).\\
- Two rooms are separated by `WALL`s, with a single `DOOR` connecting them.\\

\#\#\# Actions:\\
You can perform the following actions:\\
- `LEFT`: Turn 90° to the left.\\
- `RIGHT`: Turn 90° to the right.\\
- `MOVE`: Move one step forward in the direction you're facing.\\
- `PICKUP`: Pick up an object located in the cell *directly in front* of you.\\
- `DROP`: Drop the object you're holding into the cell *directly in front* of you.\\
- `UNLOCK`: Unlock the door located *directly in front* of you, if you are holding the `KEY`.\\

\#\#\# Rules:\\
- Before picking up a new object, you must `DROP` the one you’re holding into an **empty adjacent cell** (no `WALL`, `DOOR`, or object).\\
- You can only `PICKUP` or `UNLOCK` if:\\
  - The target cell is directly in front of you (i.e., one step in the direction you're facing).\\
  - You are adjacent to that cell.\\
- You cannot enter a cell that contains any object unless it’s been picked up.\\

\#\#\# Input:\\
You will be provided:\\
- The complete 2D grid as a list of lists (example below).\\
- Your starting facing direction.\\

Below is a sample input grid and starting direction along with a sample solution to this grid.
\begin{scriptsize}

\begin{verbatim}
<sample_grid>
[
["WALL","WALL","WALL","WALL","WALL","WALL","WALL","WALL","WALL","WALL","WALL"],
["WALL","","","","KEY","WALL","","","","","WALL"],
["WALL","","","AGENT","","WALL","","","","","WALL"],
["WALL","","","","","DOOR","","","","","WALL"],
["WALL","","","","","WALL","","","","","WALL"],
["WALL","WALL","WALL","WALL","WALL","WALL","WALL","WALL","WALL","WALL","WALL"]
]
</sample_grid>
<start_direction>
RIGHT
</start_direction>
<sample_solution>
[
  # Starting at (2, 3) facing RIGHT
  "LEFT", "MOVE", "RIGHT" # Move to (1, 3) and face RIGHT to face the KEY at (1, 4)
  "PICKUP", # Pick up the KEY at (1, 4)
  "RIGHT", "MOVE", "MOVE", "LEFT", # Move to (3, 4) and face RIGHT to face the DOOR\
  at (3, 5)
  "UNLOCK", # Unlock the DOOR at (3, 5)
]
<sample_solution>
\end{verbatim}
\end{scriptsize}
</game\_instructions>

<your\_code>\\
\# The code from the previous iteration inserted here\\
</your\_code>\\
<sample\_grids>\\
\# Grid 1 and start\_direction with worst code performance inserted here\\
--------------\\
\# Grid 2 and start\_direction with worst code performance inserted here\\
--------------\\
\# Grid 3 and start\_direction with worst code performance inserted here\\
</sample\_grids>
\end{quote}

\textbf{Direct Answer}

\begin{quote}
You are a game-playing agent in a 2D grid-based environment. Your objective is to unlock a `DOOR`.

\#\#\# Environment Description:\\
- The grid is represented as a 2D array, with the top-left corner at `[0][0]`. Each cell can contain one of the following objects:\\
  - `WALL`, `DOOR`, `KEY`, `AGENT` (You), or an empty string (`""`).\\
- You start in a specific cell facing one of four directions: `UP`, `DOWN`, `LEFT`, or `RIGHT` (this will be provided).\\
- Two rooms are separated by `WALL`s, with a single `DOOR` connecting them.\\

\#\#\# Actions:\\
You can perform the following actions:\\
- `LEFT`: Turn 90° to the left.\\
- `RIGHT`: Turn 90° to the right.\\
- `MOVE`: Move one step forward in the direction you're facing.\\
- `PICKUP`: Pick up an object located in the cell *directly in front* of you.\\
- `DROP`: Drop the object you're holding into the cell *directly in front* of you.\\
- `UNLOCK`: Unlock the door located *directly in front* of you, if you are holding the `KEY`.\\

\#\#\# Rules:\\
- Before picking up a new object, you must `DROP` the one you’re holding into an **empty adjacent cell** (no `WALL`, `DOOR`, or object).\\
- You can only `PICKUP` or `UNLOCK` if:\\
  - The target cell is directly in front of you (i.e., one step in the direction you're facing).\\
  - You are adjacent to that cell.\\
- You cannot enter a cell that contains any object unless it’s been picked up.\\

\#\#\# Input:\\
You will be provided:\\
- The complete 2D grid as a list of lists (example below).\\
- Your starting facing direction.\\

\#\#\# Output:\\
Your output should be a list of actions in the following format:\\
<actions>\\
    {["LEFT", "RIGHT", "MOVE", ...]}\\
</actions>\\
Make sure to use <actions> and </actions> XML tags for your output.

Below is a sample input grid and starting direction.

<grid>\\
\# Input grid inserted here\\
</grid>\\
<start\_direction>\\
\# Input start direction inserted here\\
</start\_direction>\\
\end{quote}

\textbf{CoT}

\begin{quote}
You are a game-playing agent in a 2D grid-based environment. Your objective is to unlock a `DOOR`.

\#\#\# Environment Description:\\
- The grid is represented as a 2D array, with the top-left corner at `[0][0]`. Each cell can contain one of the following objects:\\
  - `WALL`, `DOOR`, `KEY`, `AGENT` (You), or an empty string (`""`).\\
- You start in a specific cell facing one of four directions: `UP`, `DOWN`, `LEFT`, or `RIGHT` (this will be provided).\\
- Two rooms are separated by `WALL`s, with a single `DOOR` connecting them.\\

\#\#\# Actions:\\
You can perform the following actions:\\
- `LEFT`: Turn 90° to the left.\\
- `RIGHT`: Turn 90° to the right.\\
- `MOVE`: Move one step forward in the direction you're facing.\\
- `PICKUP`: Pick up an object located in the cell *directly in front* of you.\\
- `DROP`: Drop the object you're holding into the cell *directly in front* of you.\\
- `UNLOCK`: Unlock the door located *directly in front* of you, if you are holding the `KEY`.\\

\#\#\# Rules:\\
- Before picking up a new object, you must `DROP` the one you’re holding into an **empty adjacent cell** (no `WALL`, `DOOR`, or object).\\
- You can only `PICKUP` or `UNLOCK` if:\\
  - The target cell is directly in front of you (i.e., one step in the direction you're facing).\\
  - You are adjacent to that cell.\\
- You cannot enter a cell that contains any object unless it’s been picked up.\\

\#\#\# Input:\\
You will be provided:\\
- The complete 2D grid as a list of lists (example below).\\
- Your starting facing direction.\\

\#\#\# Output:\\
Your output should be a list of actions in the following format:\\
<actions>\\
    {["LEFT", "RIGHT", "MOVE", ...]}\\
</actions>\\
Make sure to use <actions> and </actions> XML tags for your output.

Below is a sample input grid and starting direction.

<grid>\\
\# Input grid inserted here\\
</grid>\\
<start\_direction>\\
\# Input start direction inserted here\\
</start\_direction>\\

Let's think step by step.
\end{quote}

\textbf{2-Step CoT}

1. 
\begin{quote}
You are a game-playing agent in a 2D grid-based environment. Your objective is to unlock a `DOOR`.

\#\#\# Environment Description:\\
- The grid is represented as a 2D array, with the top-left corner at `[0][0]`. Each cell can contain one of the following objects:\\
  - `WALL`, `DOOR`, `KEY`, `AGENT` (You), or an empty string (`""`).\\
- You start in a specific cell facing one of four directions: `UP`, `DOWN`, `LEFT`, or `RIGHT` (this will be provided).\\
- Two rooms are separated by `WALL`s, with a single `DOOR` connecting them.\\

\#\#\# Actions:\\
You can perform the following actions:\\
- `LEFT`: Turn 90° to the left.\\
- `RIGHT`: Turn 90° to the right.\\
- `MOVE`: Move one step forward in the direction you're facing.\\
- `PICKUP`: Pick up an object located in the cell *directly in front* of you.\\
- `DROP`: Drop the object you're holding into the cell *directly in front* of you.\\
- `UNLOCK`: Unlock the door located *directly in front* of you, if you are holding the `KEY`.\\

\#\#\# Rules:\\
- Before picking up a new object, you must `DROP` the one you’re holding into an **empty adjacent cell** (no `WALL`, `DOOR`, or object).\\
- You can only `PICKUP` or `UNLOCK` if:\\
  - The target cell is directly in front of you (i.e., one step in the direction you're facing).\\
  - You are adjacent to that cell.\\
- You cannot enter a cell that contains any object unless it’s been picked up.\\

\#\#\# Input:\\
You will be provided:\\
- The complete 2D grid as a list of lists (example below).\\
- Your starting facing direction.\\

\#\#\# Output:\\
Your output should be a list of actions in the following format:\\
<actions>\\
    {["LEFT", "RIGHT", "MOVE", ...]}\\
</actions>\\
Make sure to use <actions> and </actions> XML tags for your output.
Below is a sample input grid and starting direction.

<grid>\\
\# Input grid inserted here\\
</grid>\\
<start\_direction>\\
\# Input start direction inserted here\\
</start\_direction>\\

First provide only the actions until you pickup the key. Based on those actions, an updated grid will be given to you, after which you will have to give your entire action sequence.
\end{quote}

2.
\begin{quote}
The updated grid is:\\
<updated\_grid>\\
\# The grid after applying the initial actions is inserted here\\
</updated\_grid>\\
After those moves, you are facing [\# agent direction after the initial actions] [if holding an item: "while holding [ITEM\_NAME]"]. Now complete your actions and give your entire actions sequence.
\end{quote}

\subsubsection{MiniGrid - Door Key}\label{app:door_key_prompt}

\textbf{Direct Generation}

1.

\begin{quote}
You are a game-playing agent in a 2D grid-based environment. Your objective is to unlock a `DOOR` and reach a `GOAL` which is located in a different room than the one you start in.

\#\#\# Your Task:\\
You are **not** expected to solve the game for a specific instance.
Instead, your task is to **write pseudocode** that outlines a general strategy for solving any valid instance of this game.  
Your pseudocode should describe an algorithm that works on **any grid** and under **any valid configuration** of the objects and the agent’s starting direction. The score will be based on the number of actions taken, so the fewer actions you take, the better your score.

\#\#\# Environment Description:\\
- The grid is represented as a 2D array, with the top-left corner at `[0][0]`. Each cell can contain one of the following objects:\\
  - `WALL`, `DOOR`, `KEY`, `GOAL`, `AGENT` (You), or an empty string (`""`).\\
- You start in a specific cell facing one of four directions: `UP`, `DOWN`, `LEFT`, or `RIGHT` (this will be provided).\\
- Two rooms are separated by `WALL`s, with a single `DOOR` connecting them.\\

\#\#\# Actions:\\
You can perform the following actions:\\
- `LEFT`: Turn 90° to the left.\\
- `RIGHT`: Turn 90° to the right.\\
- `MOVE`: Move one step forward in the direction you're facing.\\
- `PICKUP`: Pick up an object located in the cell *directly in front* of you.\\
- `DROP`: Drop the object you're holding into the cell *directly in front* of you.\\
- `UNLOCK`: Unlock the door located *directly in front* of you, if you are holding the `KEY`.\\

\#\#\# Rules:\\
- You can only `PICKUP` or `UNLOCK` if:\\
  - The target cell is directly in front of you (i.e., one step in the direction you're facing).\\
  - You are adjacent to that cell.\\
- You cannot enter a cell that contains any object unless it’s been picked up.\\
- You must stand on the goal cell to complete the game.\\

\#\#\# Input:\\
You will be provided:\\
- The complete 2D grid as a list of lists (example below).\\
- Your starting facing direction.\\

Below is a sample input grid and starting direction along with a sample solution to this grid.
\begin{scriptsize}
    
\begin{verbatim}
<sample_grid>
[
["WALL","WALL","WALL","WALL","WALL","WALL","WALL","WALL"],
["WALL","","","","DOOR","","","WALL"],
["WALL","","","","WALL","","","WALL"],
["WALL","","","","WALL","","","WALL"],
["WALL","","","KEY","WALL","","","WALL"],
["WALL","AGENT","","","WALL","","","WALL"],
["WALL","","","","WALL","","GOAL","WALL"],
["WALL","WALL","WALL","WALL","WALL","WALL","WALL","WALL"]
]
</sample_grid>
<start_direction>
DOWN
</start_direction>
<sample_solution>
[
  # Starting at (5, 1) facing RIGHT
  "LEFT", "MOVE", "MOVE", "LEFT", # Move to (5, 3) facing UP to face the \
  `KEY` at (4, 3) 
  "PICKUP", # Pick up the `KEY`
  "MOVE", "MOVE", "MOVE", "MOVE", "RIGHT", # Move to (1, 3) facing RIGHT to \
  face the `DOOR` at (1, 4)
  "UNLOCK", # Unlock the `DOOR`
  "MOVE", "MOVE", "MOVE", "RIGHT", "MOVE", "MOVE", "MOVE", "MOVE", "MOVE" \
  # Move to (6, 6) which is the `GOAL` cell
]
<sample_solution>
\end{verbatim}
\end{scriptsize}

\end{quote}

2.

\begin{quote}
Now, using the pseudocode you previously wrote, translate it into Python code.
Write a function `solve(grid, start\_direction)` that returns the list of actions to complete the task.\\  
- `grid` is a 2D list representing the environment.\\
- `start\_direction` is a string: one of `"UP"`, `"DOWN"`, `"LEFT"`, or `"RIGHT"`.\\
\end{quote}

\textbf{Iterative Refinement:}

\begin{scriptsize}
    
\begin{quote}
    
You were tasked with writing code to play a game optimally. But your code didn’t perform well on certain types of grids. You are now given the entire problem statement, along with your code and some grid examples where your code doesnt perform properly. Observe and understand from the sample grids given, where your code goes wrong. Modify your code to perform well on these types of grids.  
<game\_instructions> 
You are a game-playing agent in a 2D grid-based environment. Your objective is to unlock a `DOOR` and reach a `GOAL` which is located in a different room than the one you start in.\\

\#\#\# Environment Description:\\
- The grid is represented as a 2D array, with the top-left corner at `[0][0]`. Each cell can contain one of the following objects:\\
  - `WALL`, `DOOR`, `KEY`, `GOAL`, `AGENT` (You), or an empty string (`""`).\\
- You start in a specific cell facing one of four directions: `UP`, `DOWN`, `LEFT`, or `RIGHT` (this will be provided).\\
- Two rooms are separated by `WALL`s, with a single `DOOR` connecting them.\\

\#\#\# Actions:\\
You can perform the following actions:\\
- `LEFT`: Turn 90° to the left.\\
- `RIGHT`: Turn 90° to the right.\\
- `MOVE`: Move one step forward in the direction you're facing.\\
- `PICKUP`: Pick up an object located in the cell *directly in front* of you.\\
- `DROP`: Drop the object you're holding into the cell *directly in front* of you.\\
- `UNLOCK`: Unlock the door located *directly in front* of you, if you are holding the `KEY`.\\

\#\#\# Rules:\\
- You can only `PICKUP` or `UNLOCK` if:\\
  - The target cell is directly in front of you (i.e., one step in the direction you're facing).\\
  - You are adjacent to that cell.\\
- You cannot enter a cell that contains any object unless it’s been picked up.\\
- You must stand on the goal cell to complete the game.\\

\#\#\# Input:\\
You will be provided:\\
- The complete 2D grid as a list of lists (example below).\\
- Your starting facing direction.\\

Below is a sample input grid and starting direction along with a sample solution to this grid.

\begin{verbatim}
<sample_grid>
[
["WALL","WALL","WALL","WALL","WALL","WALL","WALL","WALL"],
["WALL","","","","DOOR","","","WALL"],
["WALL","","","","WALL","","","WALL"],
["WALL","","","","WALL","","","WALL"],
["WALL","","","KEY","WALL","","","WALL"],
["WALL","AGENT","","","WALL","","","WALL"],
["WALL","","","","WALL","","GOAL","WALL"],
["WALL","WALL","WALL","WALL","WALL","WALL","WALL","WALL"]
]
</sample_grid>
<start_direction>
DOWN
</start_direction>
<sample_solution>
[
  # Starting at (5, 1) facing RIGHT
  "LEFT", "MOVE", "MOVE", "LEFT", # Move to (5, 3) facing UP to face the \
  `KEY` at (4, 3) 
  "PICKUP", # Pick up the `KEY`
  "MOVE", "MOVE", "MOVE", "MOVE", "RIGHT", # Move to (1, 3) facing RIGHT to \
  face the `DOOR` at (1, 4)
  "UNLOCK", # Unlock the `DOOR`
  "MOVE", "MOVE", "MOVE", "RIGHT", "MOVE", "MOVE", "MOVE", "MOVE", "MOVE" \
  # Move to (6, 6) which is the `GOAL` cell
]
<sample_solution>
\end{verbatim}
</game\_instructions>

<your\_code>\\
\# The code from the previous iteration inserted here\\
</your\_code>\\
<sample\_grids>\\
\# Grid 1 and start\_direction with worst code performance inserted here\\
--------------\\
\# Grid 2 and start\_direction with worst code performance inserted here\\
--------------\\
\# Grid 3 and start\_direction with worst code performance inserted here\\
</sample\_grids>
\end{quote}
\end{scriptsize}

\textbf{Direct Answer}

\begin{quote}
You are a game-playing agent in a 2D grid-based environment. Your objective is to unlock a `DOOR` and reach a `GOAL` which is located in a different room than the one you start in.

\#\#\# Environment Description:\\
- The grid is represented as a 2D array, with the top-left corner at `[0][0]`. Each cell can contain one of the following objects:\\
  - `WALL`, `DOOR`, `KEY`, `GOAL`, `AGENT` (You), or an empty string (`""`).\\
- You start in a specific cell facing one of four directions: `UP`, `DOWN`, `LEFT`, or `RIGHT` (this will be provided).\\
- Two rooms are separated by `WALL`s, with a single `DOOR` connecting them.\\

\#\#\# Actions:
You can perform the following actions:\\
- `LEFT`: Turn 90° to the left.\\
- `RIGHT`: Turn 90° to the right.\\
- `MOVE`: Move one step forward in the direction you're facing.\\
- `PICKUP`: Pick up an object located in the cell *directly in front* of you.\\
- `DROP`: Drop the object you're holding into the cell *directly in front* of you.\\
- `UNLOCK`: Unlock the door located *directly in front* of you, if you are holding the `KEY`.\\

\#\#\# Rules:\\
- You can only `PICKUP` or `UNLOCK` if:\\
  - The target cell is directly in front of you (i.e., one step in the direction you're facing).\\
  - You are adjacent to that cell.\\
- You cannot enter a cell that contains any object unless it’s been picked up.\\
- You must stand on the goal cell to complete the game.\\

\#\#\# Input:\\
You will be provided:\\
- The complete 2D grid as a list of lists (example below).\\
- Your starting facing direction.\\

\#\#\# Output:\\
Your output should be a list of actions in the following format:\\
<actions>\\
    {["LEFT", "RIGHT", "MOVE", ...]}\\
</actions>\\
Make sure to use <actions> and </actions> XML tags for your output.

Below is a sample input grid and starting direction.

<grid>\\
\# Input grid inserted here\\
</grid>\\
<start\_direction>\\
\# Input start direction inserted here\\
</start\_direction>\\
\end{quote}

\textbf{CoT}

\begin{quote}
You are a game-playing agent in a 2D grid-based environment. Your objective is to unlock a `DOOR` and reach a `GOAL` which is located in a different room than the one you start in.

\#\#\# Environment Description:\\
- The grid is represented as a 2D array, with the top-left corner at `[0][0]`. Each cell can contain one of the following objects:\\
  - `WALL`, `DOOR`, `KEY`, `GOAL`, `AGENT` (You), or an empty string (`""`).\\
- You start in a specific cell facing one of four directions: `UP`, `DOWN`, `LEFT`, or `RIGHT` (this will be provided).\\
- Two rooms are separated by `WALL`s, with a single `DOOR` connecting them.\\

\#\#\# Actions:
You can perform the following actions:\\
- `LEFT`: Turn 90° to the left.\\
- `RIGHT`: Turn 90° to the right.\\
- `MOVE`: Move one step forward in the direction you're facing.\\
- `PICKUP`: Pick up an object located in the cell *directly in front* of you.\\
- `DROP`: Drop the object you're holding into the cell *directly in front* of you.\\
- `UNLOCK`: Unlock the door located *directly in front* of you, if you are holding the `KEY`.\\

\#\#\# Rules:\\
- You can only `PICKUP` or `UNLOCK` if:\\
  - The target cell is directly in front of you (i.e., one step in the direction you're facing).\\
  - You are adjacent to that cell.\\
- You cannot enter a cell that contains any object unless it’s been picked up.\\
- You must stand on the goal cell to complete the game.\\

\#\#\# Input:\\
You will be provided:\\
- The complete 2D grid as a list of lists (example below).\\
- Your starting facing direction.\\

\#\#\# Output:\\
Your output should be a list of actions in the following format:\\
<actions>\\
    {["LEFT", "RIGHT", "MOVE", ...]}\\
</actions>\\
Make sure to use <actions> and </actions> XML tags for your output.

Below is a sample input grid and starting direction.

<grid>\\
\# Input grid inserted here\\
</grid>\\
<start\_direction>\\
\# Input start direction inserted here\\
</start\_direction>\\

Let's think step by step.
\end{quote}

\textbf{2-Step CoT}

1.
\begin{quote}
You are a game-playing agent in a 2D grid-based environment. Your objective is to unlock a `DOOR` and reach a `GOAL` which is located in a different room than the one you start in.

\#\#\# Environment Description:\\
- The grid is represented as a 2D array, with the top-left corner at `[0][0]`. Each cell can contain one of the following objects:\\
  - `WALL`, `DOOR`, `KEY`, `GOAL`, `AGENT` (You), or an empty string (`""`).\\
- You start in a specific cell facing one of four directions: `UP`, `DOWN`, `LEFT`, or `RIGHT` (this will be provided).\\
- Two rooms are separated by `WALL`s, with a single `DOOR` connecting them.\\

\#\#\# Actions:
You can perform the following actions:\\
- `LEFT`: Turn 90° to the left.\\
- `RIGHT`: Turn 90° to the right.\\
- `MOVE`: Move one step forward in the direction you're facing.\\
- `PICKUP`: Pick up an object located in the cell *directly in front* of you.\\
- `DROP`: Drop the object you're holding into the cell *directly in front* of you.\\
- `UNLOCK`: Unlock the door located *directly in front* of you, if you are holding the `KEY`.\\

\#\#\# Rules:\\
- You can only `PICKUP` or `UNLOCK` if:\\
  - The target cell is directly in front of you (i.e., one step in the direction you're facing).\\
  - You are adjacent to that cell.\\
- You cannot enter a cell that contains any object unless it’s been picked up.\\
- You must stand on the goal cell to complete the game.\\

\#\#\# Input:\\
You will be provided:\\
- The complete 2D grid as a list of lists (example below).\\
- Your starting facing direction.\\

\#\#\# Output:\\
Your output should be a list of actions in the following format:\\
<actions>\\
    {["LEFT", "RIGHT", "MOVE", ...]}\\
</actions>\\
Make sure to use <actions> and </actions> XML tags for your output.

Below is a sample input grid and starting direction.

<grid>\\
\# Input grid inserted here\\
</grid>\\
<start\_direction>\\
\# Input start direction inserted here\\
</start\_direction>\\

First provide only the actions until you unlock the door. Based on those actions, an updated grid will be given to you, after which you will have to give your entire action sequence.
\end{quote}

2.
\begin{quote}
The updated grid is:\\
<updated\_grid>\\
\# The grid after applying the initial actions is inserted here\\
</updated\_grid>\\
After those moves, you are facing [\# agent direction after the initial actions] [if holding an item: "while holding [ITEM\_NAME]"]. Now complete your actions and give your entire actions sequence.
\end{quote}

\subsubsection{MiniGrid - Unlock Pickup}\label{app:unlock_pickup_prompt}

\textbf{Direct Generation}

1.
\begin{scriptsize}
    
\begin{quote}
You are a game-playing agent in a 2D grid-based environment. Your objective is to pick up a `BOX` that is located in a different room than your starting position.\\

\#\#\# Your Task:\\
You are **not** expected to solve the game for a specific instance.  
Instead, your task is to **write pseudocode** that outlines a general strategy for solving any valid instance of this game.  
Your pseudocode should describe an algorithm that works on **any grid** and under **any valid configuration** of the objects and the agent’s starting direction. The score will be based on the number of steps taken, so the fewer steps you take, the better your score.\\

\#\#\# Environment Description:\\
- The grid is represented as a 2D array, with the top-left corner at `[0][0]`. Each cell can \\contain one of the following objects:\\
  - `WALL`, `DOOR`, `KEY`, `BOX`, `AGENT` (You), or an empty string (`""`).\\
- You start in a specific cell facing one of four directions: `UP`, `DOWN`, `LEFT`, or `RIGHT` \\(this will be provided).\\
- Two rooms are separated by `WALL`s, with a single `DOOR` connecting them.\\

\#\#\# Actions:\\
You can perform the following actions:\\
- `LEFT`: Turn 90° to the left.\\
- `RIGHT`: Turn 90° to the right.\\
- `MOVE`: Move one step forward in the direction you're facing.\\
- `PICKUP`: Pick up an object located in the cell *directly in front* of you.\\
- `DROP`: Drop the object you're holding into the cell *directly in front* of you.\\
- `UNLOCK`: Unlock the door located *directly in front* of you, if you are holding the `KEY`.\\

\#\#\# Rules:\\
- You may only hold **one object at a time** (`KEY` or `BOX`).\\
- Before picking up a new object, you must `DROP` the one you’re holding into an **empty adjacent cell** (no `WALL`, `DOOR`, or object).\\
- You can only `PICKUP` or `UNLOCK` if:\\
  - The target cell is directly in front of you (i.e., one step in the direction you're facing).\\
  - You are adjacent to that cell.\\
- You cannot:\\
  - Enter a cell that contains any object unless it’s been picked up or is a now-empty `DOOR` cell.\\
  - Drop objects into non-empty cells or into a `WALL` or `DOOR` cell.\\
  - Pick up an object (e.g. `BOX`) when you are already holding one (e.g. `KEY`).\\

\#\#\# Input:\\
You will be provided:\\
- The complete 2D grid as a list of lists (example below).\\
- Your starting facing direction.\\

\begin{verbatim}
<sample_grid>
<grid>
[["WALL","WALL","WALL","WALL","WALL","WALL","WALL","WALL","WALL","WALL","WALL"],
["WALL","","","","","DOOR","","","","","WALL"],
["WALL","","","","","WALL","","","","","WALL"],
["WALL","","AGENT","","KEY","WALL","","","","","WALL"],
["WALL","","","","","WALL","","","","BOX","WALL"],
["WALL","WALL","WALL","WALL","WALL","WALL","WALL","WALL","WALL","WALL","WALL"]]
</grid>
<start_direction>
DOWN
</start_direction>

<sample_solution>
[
  # Starting at (3,2) facing DOWN
  "LEFT", "MOVE", # Move to (3, 3) and face RIGHT to face the KEY at (3, 4)
  "PICKUP", # Pick up the KEY at (3, 4)
  "LEFT", "MOVE", "MOVE", "RIGHT", "MOVE", # Move to (1, 4) and face RIGHT \
  to face the DOOR at (1, 5)
  "UNLOCK", # Unlock the DOOR at (1, 5)
  "MOVE", "MOVE", "RIGHT", "MOVE", "MOVE", "MOVE", "LEFT", "MOVE", "MOVE", \
  # Move to (4,8) near the BOX at (4, 9)
  "LEFT", "DROP", # Drop the KEY on the empty cell at (3, 8) before picking \
  up the BOX
  "RIGHT", # Face RIGHT at (4, 8) to face the BOX at (4, 9)
  "PICKUP", # Pick up the BOX at (4, 9)
]
</sample_solution>
</sample_grid>
\end{verbatim}
\end{quote}
\end{scriptsize}

2.
\begin{quote}
Now, using the pseudocode you previously wrote, translate it into Python code.

Write a function `solve(grid, start\_direction)` that returns the list of actions to complete the task.  
- `grid` is a 2D list representing the environment.
- `start\_direction` is a string: one of `"UP"`, `"DOWN"`, `"LEFT"`, or `"RIGHT"`.
\end{quote}

\textbf{Iterative Refinement}

\begin{scriptsize}
    
\begin{quote}
You were tasked with writing code to play a game optimally. But your code didn’t perform well on certain types of grids. You are now given the entire problem statement, along with your code and some grid examples where your code doesnt perform properly. Observe and understand from the sample grids given, where your code goes wrong. Modify your code to perform well on these types of grids.

<game\_instructions>  
You are a game-playing agent in a 2D grid-based environment. Your objective is to pick up a `BOX` that is located in a different room than your starting position.\\

\#\#\# Environment Description:\\
- The grid is represented as a 2D array, with the top-left corner at `[0][0]`. Each cell can \\contain one of the following objects:\\
  - `WALL`, `DOOR`, `KEY`, `BOX`, `AGENT` (You), or an empty string (`""`).\\
- You start in a specific cell facing one of four directions: `UP`, `DOWN`, `LEFT`, or `RIGHT` \\(this will be provided).\\
- Two rooms are separated by `WALL`s, with a single `DOOR` connecting them.\\

\#\#\# Actions:\\
You can perform the following actions:\\
- `LEFT`: Turn 90° to the left.\\
- `RIGHT`: Turn 90° to the right.\\
- `MOVE`: Move one step forward in the direction you're facing.\\
- `PICKUP`: Pick up an object located in the cell *directly in front* of you.\\
- `DROP`: Drop the object you're holding into the cell *directly in front* of you.\\
- `UNLOCK`: Unlock the door located *directly in front* of you, if you are holding the `KEY`.\\

\#\#\# Rules:\\
- You may only hold **one object at a time** (`KEY` or `BOX`).\\
- Before picking up a new object, you must `DROP` the one you’re holding into an **empty adjacent cell** (no `WALL`, `DOOR`, or object).\\
- You can only `PICKUP` or `UNLOCK` if:\\
  - The target cell is directly in front of you (i.e., one step in the direction you're facing).\\
  - You are adjacent to that cell.\\
- You cannot:\\
  - Enter a cell that contains any object unless it’s been picked up or is a now-empty `DOOR` cell.\\
  - Drop objects into non-empty cells or into a `WALL` or `DOOR` cell.\\
  - Pick up an object (e.g. `BOX`) when you are already holding one (e.g. `KEY`).\\

\#\#\# Input:\\
You will be provided:\\
- The complete 2D grid as a list of lists (example below).\\
- Your starting facing direction.\\

\begin{verbatim}
<sample_grid>
<grid>
[["WALL","WALL","WALL","WALL","WALL","WALL","WALL","WALL","WALL","WALL","WALL"],
["WALL","","","","","DOOR","","","","","WALL"],
["WALL","","","","","WALL","","","","","WALL"],
["WALL","","AGENT","","KEY","WALL","","","","","WALL"],
["WALL","","","","","WALL","","","","BOX","WALL"],
["WALL","WALL","WALL","WALL","WALL","WALL","WALL","WALL","WALL","WALL","WALL"]]
</grid>
<start_direction>
DOWN
</start_direction>

<sample_solution>
[
  # Starting at (3,2) facing DOWN
  "LEFT", "MOVE", # Move to (3, 3) and face RIGHT to face the KEY at (3, 4)
  "PICKUP", # Pick up the KEY at (3, 4)
  "LEFT", "MOVE", "MOVE", "RIGHT", "MOVE", # Move to (1, 4) and face RIGHT \
  to face the DOOR at (1, 5)
  "UNLOCK", # Unlock the DOOR at (1, 5)
  "MOVE", "MOVE", "RIGHT", "MOVE", "MOVE", "MOVE", "LEFT", "MOVE", "MOVE", \
  # Move to (4,8) near the BOX at (4, 9)
  "LEFT", "DROP", # Drop the KEY on the empty cell at (3, 8) before picking \
  up the BOX
  "RIGHT", # Face RIGHT at (4, 8) to face the BOX at (4, 9)
  "PICKUP", # Pick up the BOX at (4, 9)
]
</sample_solution>
</sample_grid>
\end{verbatim}
</game\_instructions>

<your\_code>\\
\# The code from the previous iteration inserted here\\
</your\_code>\\
<sample\_grids>\\
\# Grid 1 and start\_direction with worst code performance inserted here\\
--------------\\
\# Grid 2 and start\_direction with worst code performance inserted here\\
--------------\\
\# Grid 3 and start\_direction with worst code performance inserted here\\
</sample\_grids>
\end{quote}
\end{scriptsize}

\textbf{Direct Answer}
\begin{quote}
You are a game-playing agent in a 2D grid-based environment. Your objective is to pick up a `BOX` that is located in a different room than your starting position.

\#\#\# Environment Description:\\
- The grid is represented as a 2D array, with the top-left corner at `[0][0]`. Each cell can contain one of the following objects:\\
  - `WALL`, `DOOR`, `KEY`, `BOX`, `AGENT` (You), or an empty string (`""`).\\
- You start in a specific cell facing one of four directions: `UP`, `DOWN`, `LEFT`, or `RIGHT` (this will be provided).\\
- Two rooms are separated by `WALL`s, with a single `DOOR` connecting them.\\

\#\#\# Actions:\\
You can perform the following actions:\\
- `LEFT`: Turn 90° to the left.\\
- `RIGHT`: Turn 90° to the right.\\
- `MOVE`: Move one step forward in the direction you're facing.\\
- `PICKUP`: Pick up an object located in the cell *directly in front* of you.\\
- `DROP`: Drop the object you're holding into the cell *directly in front* of you.\\
- `UNLOCK`: Unlock the door located *directly in front* of you, if you are holding the `KEY`.\\

\#\#\# Rules:\\
- You may only hold **one object at a time** (`KEY` or `BOX`).\\
- Before picking up a new object, you must `DROP` the one you’re holding into an **empty adjacent cell** (no `WALL`, `DOOR`, or object).\\
- You can only `PICKUP` or `UNLOCK` if:\\
  - The target cell is directly in front of you (i.e., one step in the direction you're facing).\\
  - You are adjacent to that cell.\\
- You cannot:\\
  - Enter a cell that contains any object unless it’s been picked up or is a now-empty `DOOR` cell.\\
  - Drop objects into non-empty cells or into a `WALL` or `DOOR` cell.\\
  - Pick up an object (e.g. `BOX`) when you are already holding one (e.g. `KEY`).\\

\#\#\# Input:\\
You will be provided:\\
- The complete 2D grid as a list of lists (example below).\\
- Your starting facing direction.\\

\#\#\# Output:\\
Your output should be a list of actions in the following format:\\
<actions>\\
    {["LEFT", "RIGHT", "MOVE", ...]}\\
</actions>\\
Make sure to use <actions> and </actions> XML tags for your output.

Below is a sample input grid and starting direction.

<grid>\\
\# Input grid inserted here\\
</grid>\\
<start\_direction>\\
\# Input start direction inserted here\\
</start\_direction>\\
\end{quote}

\textbf{CoT}
\begin{quote}
You are a game-playing agent in a 2D grid-based environment. Your objective is to pick up a `BOX` that is located in a different room than your starting position.

\#\#\# Environment Description:\\
- The grid is represented as a 2D array, with the top-left corner at `[0][0]`. Each cell can contain one of the following objects:\\
  - `WALL`, `DOOR`, `KEY`, `BOX`, `AGENT` (You), or an empty string (`""`).\\
- You start in a specific cell facing one of four directions: `UP`, `DOWN`, `LEFT`, or `RIGHT` (this will be provided).\\
- Two rooms are separated by `WALL`s, with a single `DOOR` connecting them.\\

\#\#\# Actions:\\
You can perform the following actions:\\
- `LEFT`: Turn 90° to the left.\\
- `RIGHT`: Turn 90° to the right.\\
- `MOVE`: Move one step forward in the direction you're facing.\\
- `PICKUP`: Pick up an object located in the cell *directly in front* of you.\\
- `DROP`: Drop the object you're holding into the cell *directly in front* of you.\\
- `UNLOCK`: Unlock the door located *directly in front* of you, if you are holding the `KEY`.\\

\#\#\# Rules:\\
- You may only hold **one object at a time** (`KEY` or `BOX`).\\
- Before picking up a new object, you must `DROP` the one you’re holding into an **empty adjacent cell** (no `WALL`, `DOOR`, or object).\\
- You can only `PICKUP` or `UNLOCK` if:\\
  - The target cell is directly in front of you (i.e., one step in the direction you're facing).\\
  - You are adjacent to that cell.\\
- You cannot:\\
  - Enter a cell that contains any object unless it’s been picked up or is a now-empty `DOOR` cell.\\
  - Drop objects into non-empty cells or into a `WALL` or `DOOR` cell.\\
  - Pick up an object (e.g. `BOX`) when you are already holding one (e.g. `KEY`).\\

\#\#\# Input:\\
You will be provided:\\
- The complete 2D grid as a list of lists (example below).\\
- Your starting facing direction.\\

\#\#\# Output:\\
Your output should be a list of actions in the following format:\\
<actions>\\
    {["LEFT", "RIGHT", "MOVE", ...]}\\
</actions>\\
Make sure to use <actions> and </actions> XML tags for your output.

Below is a sample input grid and starting direction.

<grid>\\
\# Input grid inserted here\\
</grid>\\
<start\_direction>\\
\# Input start direction inserted here\\
</start\_direction>\\

Let's think step by step.
\end{quote}

\textbf{2-Step CoT}

1. 
\begin{quote}
You are a game-playing agent in a 2D grid-based environment. Your objective is to pick up a `BOX` that is located in a different room than your starting position.

\#\#\# Environment Description:\\
- The grid is represented as a 2D array, with the top-left corner at `[0][0]`. Each cell can contain one of the following objects:\\
  - `WALL`, `DOOR`, `KEY`, `BOX`, `AGENT` (You), or an empty string (`""`).\\
- You start in a specific cell facing one of four directions: `UP`, `DOWN`, `LEFT`, or `RIGHT` (this will be provided).\\
- Two rooms are separated by `WALL`s, with a single `DOOR` connecting them.\\

\#\#\# Actions:\\
You can perform the following actions:\\
- `LEFT`: Turn 90° to the left.\\
- `RIGHT`: Turn 90° to the right.\\
- `MOVE`: Move one step forward in the direction you're facing.\\
- `PICKUP`: Pick up an object located in the cell *directly in front* of you.\\
- `DROP`: Drop the object you're holding into the cell *directly in front* of you.\\
- `UNLOCK`: Unlock the door located *directly in front* of you, if you are holding the `KEY`.\\

\#\#\# Rules:\\
- You may only hold **one object at a time** (`KEY` or `BOX`).\\
- Before picking up a new object, you must `DROP` the one you’re holding into an **empty adjacent cell** (no `WALL`, `DOOR`, or object).\\
- You can only `PICKUP` or `UNLOCK` if:\\
  - The target cell is directly in front of you (i.e., one step in the direction you're facing).\\
  - You are adjacent to that cell.\\
- You cannot:\\
  - Enter a cell that contains any object unless it’s been picked up or is a now-empty `DOOR` cell.\\
  - Drop objects into non-empty cells or into a `WALL` or `DOOR` cell.\\
  - Pick up an object (e.g. `BOX`) when you are already holding one (e.g. `KEY`).\\

\#\#\# Input:\\
You will be provided:\\
- The complete 2D grid as a list of lists (example below).\\
- Your starting facing direction.\\

\#\#\# Output:\\
Your output should be a list of actions in the following format:\\
<actions>\\
    {["LEFT", "RIGHT", "MOVE", ...]}\\
</actions>\\
Make sure to use <actions> and </actions> XML tags for your output.

Below is a sample input grid and starting direction.

<grid>\\
\# Input grid inserted here\\
</grid>\\
<start\_direction>\\
\# Input start direction inserted here\\
</start\_direction>\\

First provide only the actions until you unlock the door. Based on those actions, an updated grid will be given to you, after which you will have to give your entire action sequence.
\end{quote}

2.
\begin{quote}
The updated grid is:\\
<updated\_grid>\\
\# The grid after applying the initial actions is inserted here\\
</updated\_grid>\\
After those moves, you are facing [\# agent direction after the initial actions] [if holding an item: "while holding [ITEM\_NAME]"]. Now complete your actions and give your entire actions sequence.
\end{quote}

\subsection{Baseline Pseudocodes}\label{app:baseline_code}

\subsubsection{GRASP}\label{app:grasp_code}

\textbf{Random}

\begin{scriptsize}
    
\begin{verbatim}
FUNCTION RANDOM_WALK(agent, directions_num):
    # Determine the possible directions based on the number of directions allowed
    IF directions_num == 4:
        directions = ["LEFT", "RIGHT", "UP", "DOWN"]
    ELSE:
        directions = ["LEFT", "RIGHT", "UP", "DOWN", "UPLEFT", \
        "UPRIGHT", "DOWNLEFT", "DOWNRIGHT"]

    actions_taken = []       # List to keep track of actions taken for backtracking
    all_action = []          # List to store the full sequence of actions

    # Move the agent in a random direction 6 times and collect energy
    FOR i FROM 1 TO 6:
        direction = RANDOM_CHOICE(directions)
        MOVE_AGENT(agent, direction)
        actions_taken.APPEND(direction)
        all_action.APPEND(direction)
        TAKE_ENERGY(agent)
        all_action.APPEND('take')

    # Backtrack by reversing the path taken
    FOR direction IN REVERSE(actions_taken):
        opposite_direction = GET_OPPOSITE_DIRECTION(direction)
        MOVE_AGENT(agent, opposite_direction)
        all_action.APPEND(opposite_direction)

    # Drop all collected energy
    DROP_ALL_ENERGY(agent)
    all_action.APPEND('drop')

    RETURN all_action

\end{verbatim}
\end{scriptsize}

\textbf{Greedy}

\begin{scriptsize}
    
\begin{verbatim}
# This greedy algorithm doesn't include considerations for cost_per_step and carry_limit  
FUNCTION GREEDY_SOLVE(grid, start_position, is_diagonals_allowed, max_actions):
    actions = []
    current_position = start_position
    actions_remaining = max_actions
    while actions_remaining > 0:
        # Perform Breadth First Search (BFS) to find the nearest energy token
        nearest_energy_position, path = find_nearest_energy_bfs(
            grid, current_position, is_diagonals_allowed
        )
        if nearest_energy_position is None:
            # No more energy tokens available in the grid
            break
        path_length = len(path)
        # Find the number of steps required to return to the start
        # by backtracking in the same path to reach here
        steps_needed_to_return = len(
            [action for action in actions if action not in ['TAKE', 'DROP']]
        )
        # If there is not enough steps to go to the nearest energy token
        # and come back to the start, return to start and end the trip
        # The +2 is for the TAKE and DROP to be performed
        if 2*path_length + steps_needed_to_return + 2 > actions_remaining:
            # Actions to backtrack the same path taken to reach here
            return_actions = return_to_start(actions)
            actions.extend(return_actions)
            actions.append('DROP')
            return actions
        # The remaining steps allow going to the nearest energy token before
        # going back to start
        for direction in path:
            if actions_remaining <= 0:
                break
            move_agent(direction)
            actions.append(direction)
            actions_remaining -= 1
        actions.append('TAKE')
        actions_remaining -= 1
    return actions  
\end{verbatim}
\end{scriptsize}

\subsubsection{MiniGrid - Unlock}\label{app:unlock_code}
\textbf{Random}

\begin{scriptsize}
    
\begin{verbatim}
FUNCTION SOLVE(grid, start_direction):
    CONSTANT ALLOWED_ACTIONS = ["LEFT", "RIGHT", "MOVE", "PICKUP", "DROP", "UNLOCK"]
    CONSTANT NUM_MOVES = 100
    # Randomly select NUM_MOVES actions from ALLOWED_ACTIONS
    actions = RANDOM_CHOICE(ALLOWED_ACTIONS, NUM_MOVES)
    
    # Convert the actions to a list and return
    RETURN actions
\end{verbatim}
\end{scriptsize}

\textbf{Greedy}
\begin{verbatim}
FUNCTION SOLVE(grid, start_direction):
    direction = start_direction
    curr_pos = FIND_OBJECT(grid, "AGENT")

    # Locate the key in the grid
    key_pos = FIND_OBJECT(grid, "KEY")

    # Get the path to the key using BFS
    path_to_key = PATH_TO_OBJECT(grid, curr_pos, key_pos)
    
    # Add the movement and direction changes to the action list
    action_list = GET_PATH_ACTIONS(path_to_key, direction)

    # Pick up the key
    action_list.APPEND("PICKUP")

    # Locate the door in the grid
    door_pos = FIND_OBJECT(grid, "DOOR")
    
    # Get the path to the door using BFS
    path_to_door = PATH_TO_OBJECT(grid, curr_pos, door_pos)
    
    # Add the movement and direction changes to the action list
    action_list_door = GET_PATH_ACTIONS(path_to_door, direction)

    action_list.EXTEND(action_list_door)

    # Unlock the door
    action_list.APPEND("UNLOCK")

    RETURN action_list
\end{verbatim}

\subsubsection{MiniGrid - Door Key}\label{app:doorkey_code}
\textbf{Random}
\begin{scriptsize}
    
\begin{verbatim}
FUNCTION SOLVE(grid, start_direction):
    CONSTANT ALLOWED_ACTIONS = ["LEFT", "RIGHT", "MOVE", "PICKUP", "DROP", "UNLOCK"]
    CONSTANT NUM_MOVES = 100
    # Randomly select NUM_MOVES actions from ALLOWED_ACTIONS
    actions = RANDOM_CHOICE(ALLOWED_ACTIONS, NUM_MOVES)
    
    # Convert the actions to a list and return
    RETURN actions
\end{verbatim}
\end{scriptsize}

\textbf{Greedy}
\begin{verbatim}
FUNCTION SOLVE(grid, start_direction):
    direction = start_direction
    curr_pos = FIND_OBJECT(grid, "AGENT")

    key_pos = FIND_OBJECT(grid, "KEY")
    
    # Get the path to the key using BFS
    path_to_key = PATH_TO_OBJECT(grid, curr_pos, key_pos)

    action_list = GET_PATH_ACTIONS(path_to_key, direction)

    action_list.APPEND("PICKUP")

    door_pos = FIND_OBJECT(grid, "DOOR")

    path_to_door = PATH_TO_OBJECT(grid, curr_pos, door_pos)

    # Get the path to the door using BFS
    action_list_door = GET_PATH_ACTIONS(path_to_door, direction)
    action_list_door.REMOVE_LAST()

    action_list.EXTEND(action_list_door)
    action_list.APPEND("UNLOCK")

    goal_pos = FIND_OBJECT(grid, "GOAL")

    # Get the path to the goal using BFS
    path_to_goal = PATH_TO_OBJECT(grid, curr_pos, goal_pos)

    action_list_goal = GET_PATH_ACTIONS(path_to_goal, direction)

    action_list.EXTEND(action_list_goal)

    RETURN action_list
\end{verbatim}

\subsubsection{MiniGrid - Unlock Pickup}\label{app:unlockpickup_code}
\textbf{Random}
\begin{scriptsize}
    
\begin{verbatim}
FUNCTION SOLVE(grid, start_direction):
    CONSTANT ALLOWED_ACTIONS = ["LEFT", "RIGHT", "MOVE", "PICKUP", "DROP", "UNLOCK"]
    CONSTANT NUM_MOVES = 100
    # Randomly select NUM_MOVES actions from ALLOWED_ACTIONS
    actions = RANDOM_CHOICE(ALLOWED_ACTIONS, NUM_MOVES)
    
    # Convert the actions to a list and return
    RETURN actions
\end{verbatim}
\end{scriptsize}

\textbf{Greedy}
\begin{scriptsize}
    
\begin{verbatim}
FUNCTION SOLVE(grid, start_direction):
    direction = start_direction
    curr_pos = FIND_OBJECT(grid, "AGENT")

    key_pos = FIND_OBJECT(grid, "KEY")

    # Get the path to the key using BFS
    path_to_key = PATH_TO_OBJECT(grid, curr_pos, key_pos)

    action_list = GET_PATH_ACTIONS(path_to_key, direction)

    action_list.APPEND("PICKUP")

    door_pos = FIND_OBJECT(grid, "DOOR")

    # Get the path to the door using BFS
    path_to_door = PATH_TO_OBJECT(grid, curr_pos, door_pos)

    action_list_door GET_PATH_ACTIONS(path_to_door, direction)

    action_list.EXTEND(action_list_door)
    action_list.APPEND("UNLOCK")

    box_pos = FIND_OBJECT(grid, "BOX")

    # Get the path to the box using BFS
    path_to_box = PATH_TO_OBJECT(grid, curr_pos, box_pos)

    action_list_goal = ADD_DIRECTION_CHANGES(path_to_box, direction)

    action_list.EXTEND(action_list_goal)

    # Drop the key on the cell in the path we arrived
    IF action_list[LAST] EQUALS "MOVE":
        action_list.EXTEND(["RIGHT", "RIGHT", "DROP", "RIGHT", "RIGHT", "PICKUP"])
    ELSE:
        IF action_list[LAST] EQUALS "RIGHT":
            action_list.EXTEND(["RIGHT", "DROP", "LEFT", "PICKUP"])
        ELSE IF action_list[LAST] EQUALS "LEFT":
            action_list.EXTEND(["LEFT", "DROP", "RIGHT", "PICKUP"])

    RETURN action_list
\end{verbatim}
\end{scriptsize}

\newpage
\section*{NeurIPS Paper Checklist}

\begin{enumerate}

\item {\bf Claims}
    \item[] Question: Do the main claims made in the abstract and introduction accurately reflect the paper's contributions and scope?
    \item[] Answer: \answerYes{} 
    \item[] Justification: The abstract and introduction clearly state the main contributions of the paper. There is a dedicated bullet point in the introduction for contributions, with further details throughout the paper.

\item {\bf Limitations}
    \item[] Question: Does the paper discuss the limitations of the work performed by the authors?
    \item[] Answer: \answerYes{} 
    \item[] Justification: There is a section included specifically for discussion and limitations.

\item {\bf Theory assumptions and proofs}
    \item[] Question: For each theoretical result, does the paper provide the full set of assumptions and a complete (and correct) proof?
    \item[] Answer: \answerNA{} 
    \item[] Justification: \answerNA{}

    \item {\bf Experimental result reproducibility}
    \item[] Question: Does the paper fully disclose all the information needed to reproduce the main experimental results of the paper to the extent that it affects the main claims and/or conclusions of the paper (regardless of whether the code and data are provided or not)?
    \item[] Answer: \answerYes{} 
    \item[] Justification: We include all prompts used and code used in the Appendix and share an external URL to the code.

\item {\bf Open access to data and code}
    \item[] Question: Does the paper provide open access to the data and code, with sufficient instructions to faithfully reproduce the main experimental results, as described in supplemental material?
    \item[] Answer: \answerYes{} 
    \item[] Justification: We include all prompts used and code used in the Appendix and share an external URL to the code.

\item {\bf Experimental setting/details}
    \item[] Question: Does the paper specify all the training and test details (e.g., data splits, hyperparameters, how they were chosen, type of optimizer, etc.) necessary to understand the results?
    \item[] Answer: \answerYes{} 
    \item[] Justification: We detail these in the experimental setup section.

\item {\bf Experiment statistical significance}
    \item[] Question: Does the paper report error bars suitably and correctly defined or other appropriate information about the statistical significance of the experiments?
    \item[] Answer: \answerYes{} 
    \item[] Justification: Measures of statistical bound are provided with the main results where applicable.

\item {\bf Experiments compute resources}
    \item[] Question: For each experiment, does the paper provide sufficient information on the computer resources (type of compute workers, memory, time of execution) needed to reproduce the experiments?
    \item[] Answer: \answerYes{} 
    \item[] Justification: Cost of LLM prompting is described.
    
\item {\bf Code of ethics}
    \item[] Question: Does the research conducted in the paper conform, in every respect, with the NeurIPS Code of Ethics \url{https://neurips.cc/public/EthicsGuidelines}?
    \item[] Answer: \answerYes{} 
    \item[] Justification: The research conforms to the NeurIPS Code of Ethics.

\item {\bf Broader impacts}
    \item[] Question: Does the paper discuss both potential positive societal impacts and negative societal impacts of the work performed?
    \item[] Answer: \answerYes{} 
    \item[] Justification: Details are provided in the discussion and limitation section.
    
\item {\bf Safeguards}
    \item[] Question: Does the paper describe safeguards that have been put in place for responsible release of data or models that have a high risk for misuse (e.g., pretrained language models, image generators, or scraped datasets)?
    \item[] Answer: \answerNA{} 
    \item[] Justification: \answerNA{}

\item {\bf Licenses for existing assets}
    \item[] Question: Are the creators or original owners of assets (e.g., code, data, models), used in the paper, properly credited and are the license and terms of use explicitly mentioned and properly respected?
    \item[] Answer: \answerYes{} 
    \item[] Justification: The two benchmarks and six LLM models are properly cited.

\item {\bf New assets}
    \item[] Question: Are new assets introduced in the paper well documented and is the documentation provided alongside the assets?
    \item[] Answer: \answerNA{} 
    \item[] Justification: \answerNA{}

\item {\bf Crowdsourcing and research with human subjects}
    \item[] Question: For crowdsourcing experiments and research with human subjects, does the paper include the full text of instructions given to participants and screenshots, if applicable, as well as details about compensation (if any)? 
    \item[] Answer: \answerNA{} 
    \item[] Justification: \answerNA{}

\item {\bf Institutional review board (IRB) approvals or equivalent for research with human subjects}
    \item[] Question: Does the paper describe potential risks incurred by study participants, whether such risks were disclosed to the subjects, and whether Institutional Review Board (IRB) approvals (or an equivalent approval/review based on the requirements of your country or institution) were obtained?
    \item[] Answer: \answerNA{} 
    \item[] Justification: \answerNA{}

\item {\bf Declaration of LLM usage}
    \item[] Question: Does the paper describe the usage of LLMs if it is an important, original, or non-standard component of the core methods in this research? Note that if the LLM is used only for writing, editing, or formatting purposes and does not impact the core methodology, scientific rigorousness, or originality of the research, declaration is not required.
    \item[] Answer: \answerYes{} 
    \item[] Justification: LLMs usage are detailed in the experimental setup section.

\end{enumerate}

\end{document}